\pdfoutput=1

\documentclass{article} %
\usepackage{iclr2027_conference,times}

\usepackage{amsmath,amsfonts,bm}

\def\eqref#1{equation~\ref{#1}}

\def\1{\bm{1}}

\DeclareMathAlphabet{\mathsfit}{\encodingdefault}{\sfdefault}{m}{sl}
\SetMathAlphabet{\mathsfit}{bold}{\encodingdefault}{\sfdefault}{bx}{n}

\usepackage{hyperref}
\usepackage{url}

\usepackage{booktabs}
\usepackage{multirow}
\usepackage{placeins}
\usepackage{graphicx}
\usepackage{subcaption}
\usepackage{makecell}
\usepackage{xspace}
\usepackage[table]{xcolor}
\usepackage{amssymb}
\usepackage{tcolorbox}
\tcbuselibrary{breakable,skins}
\title{Scaling Long-Form Story Generation via\\Narrative State Tracking}

\author{Zhennan Wan, Jianfei Chen\thanks{Corresponding Author} \\
Dept. of Comp. Sci. and Tech., Institute for AI, 
BNRist Center, THBI Lab, \\Tsinghua-Bosch Joint ML Center, 
Tsinghua University\\
\texttt{wanzn26@mails.tsinghua.edu.cn, jianfeic@tsinghua.edu.cn}
}

\newcommand{\NstAgent}{\textsc{NstAgent}\xspace}
\definecolor{nstbg}{HTML}{E8F0FB}
\newcommand{\oursbg}{\cellcolor{nstbg}}
\definecolor{nstframe}{HTML}{A9C4EB}
\newtcolorbox{promptbox}[2][]{enhanced, #1, colback=nstbg!45, colframe=nstframe, boxrule=0.5pt, arc=2pt, left=5pt, right=5pt, top=4pt, bottom=4pt, fonttitle=\bfseries\footnotesize, coltitle=black, colbacktitle=nstbg, title={#2}, fontupper=\footnotesize, before upper={\setlength{\parskip}{3pt}\setlength{\parindent}{0pt}}}

\iclrfinalcopy %
\begin{document}

\maketitle

\begin{abstract}
LLMs have demonstrated strong capabilities in creative writing. However, scaling them to full-length novels remains challenging, as maintaining narrative consistency becomes increasingly difficult. Existing story-generation methods typically focus on stories of up to about ten thousand words, leaving their ability to scale to full-length novels underexplored.
In this work, we introduce Narrative State Tracking Agent (\NstAgent), a training-free agentic framework that allows LLMs to track a structured narrative state including characters, past events and future requirements. We extend an existing benchmark to compare narrative consistency across lengths, and use it together with a writing-quality benchmark to systematically evaluate stories ranging from 10K to 100K words. 
We show that \NstAgent achieves better narrative consistency and writing quality as stories grow longer, and neither of them degrades noticeably as length increases, suggesting that it provides an effective approach to scaling story generation toward full-length novels.
\footnote{Code and data are available at \url{https://github.com/zhennan1/NstAgent}.} 
\end{abstract}

\begin{figure}[h]
    \centering
    \begin{subfigure}[b]{0.48\textwidth}
    \includegraphics[width=\linewidth]{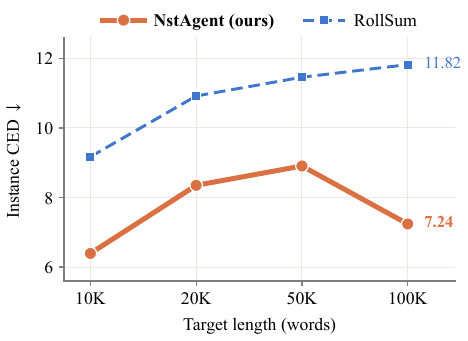} %
    \end{subfigure}
    \begin{subfigure}[b]{0.48\textwidth}
    \includegraphics[width=\linewidth]{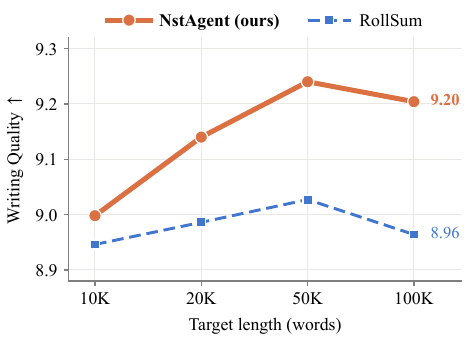} %
    \end{subfigure}
    \caption{Instance CED (left, lower is better) and writing quality (right, higher is better) of \NstAgent and RollSum on DeepSeek-V4-Flash as the target length grows from 10K to 100K words.}
    \label{fig:intro}
\end{figure}

\section{Introduction}

Recently, the output length of LLMs has grown from a few thousand words to tens of thousands~\citep{openai2025gpt5,anthropic2025claude4,qwen2025qwen3}. However, writing a novel is not the same as simply emitting a long output. As a story grows longer, its characters, events, and foreshadowings accumulate roughly linearly, yet every newly generated passage must remain compatible with \emph{all} previously established constraints. The burden of maintaining consistency therefore grows superlinearly with length and becomes the bottleneck of long-form story writing. We refer to this as the \emph{length scaling} problem of long-form story generation, where scaling denotes the extension of output length.

A straightforward approach is to train longer-writing models to extend the effective writing length~\citep{bai2025longwriter,wu2026longwriterzero}, but high-quality writing data at still greater lengths is extremely scarce and provides little effective supervision, and the training cost increases quadratically, so the marginal cost of scaling further is high. To avoid this, more works turn to hierarchical generation, and most such works decompose writing into two stages: (i) generating a story outline, and (ii) generating the story chapter by chapter conditioned on the outline~\citep{yang2022re3,yang2023doc,mirowski2023dramatron}. An outline supplies global structure, but it is static and coarse-grained: concrete settings that emerge during chapter generation cannot constrain subsequent chapters. Memory-based methods reuse the preceding text, whether by retrieval~\citep{zhou2023recurrentgpt}, by compression~\citep{xia2025storywriter}, or by feeding the full history back into the model~\citep{wu2026superwriter}; others resort to knowledge graphs~\citep{wang2025dome,li2025storyteller}, which are costly to construct and maintain. We discuss these approaches further in Section~\ref{sec:challenge}.

Therefore, we study the question: \textbf{how can long-form story generation be scaled?} We argue that a viable solution must satisfy three requirements at once: (i) the generation process is reliable and its total cost grows approximately linearly with length; (ii) narrative consistency does not degrade appreciably as length grows; and (iii) writing quality is not sacrificed to these constraints. Answering this question presupposes an evaluation protocol that is comparable across lengths. For writing quality, we adopt WritingBench~\citep{wu2025writingbench}. For narrative consistency, we build on ConStory-Bench~\citep{li2026constory}, which mainly targets stories of up to 10K words. On much longer stories, the number of errors a judge reports does not grow in proportion to length, so this density underestimates errors and cannot be compared across lengths. We revise this benchmark so that narrative consistency can be evaluated at substantially longer scales and compared across different lengths.

A long-standing view in narratology holds that a narrative text records the transition of a system from one state to another~\citep{todorov1977poetics}. Building on this view, as well as a systematic analysis of the limitations of previous methods, we propose \textbf{\NstAgent}, a training-free agentic framework that formulates long-form story generation as controlled, autonomous transitions over an explicit narrative state. Unlike free-form textual memory, the state maintained by \NstAgent is structured, typed, compact, and comprises three components: character states, events that have already occurred, and pending narrative promises, including foreshadowings, suspense, and explicit commitments. For each chapter, \NstAgent generates the prose conditioned on the current state and then immediately updates the narrative state, thereby tracking the narrative state throughout generation and better preserving narrative consistency.

In summary, our main contributions are as follows:

\begin{itemize}
    \item We characterize long-form story generation as a scaling problem along output length, and extend an existing narrative consistency benchmark so that it covers stories of up to 100K words and remains comparable across lengths.
    \item We propose \NstAgent, a training-free agentic framework that maintains narrative consistency by explicitly tracking narrative state including characters, past events and future requirements, with a total cost that grows approximately linearly with length.
    \item We conduct a systematic evaluation across multiple LLMs over the 10K--100K word range. Experiments show that \NstAgent achieves better narrative consistency and writing quality as stories grow longer, and neither of them degrades noticeably as length increases.
\end{itemize}

\section{Related Work}
\label{sec:related-work}

\paragraph{Training longer-writing models.}
LongWriter~\citep{bai2025longwriter} and Self-Lengthen~\citep{quan2024selflengthen} fine-tune models on synthesized long outputs, and LongWriter-Zero~\citep{wu2026longwriterzero} use reinforcement learning to extend what a single call can write. The whole story must still fit into one generation, long outputs remain volatile in length~\citep{he2026stablelongform}, and supervision for novel-length text is scarce. \NstAgent is complementary to these methods and requires no training.

\paragraph{Hierarchical generation.}
Plan-then-write pipelines date back to hierarchical story generation~\citep{fan2018hierarchical,yao2019planwrite} and outline-conditioned generation with plot states~\citep{rashkin2020plotmachines}. With LLMs, Re$^3$~\citep{yang2022re3} and DOC~\citep{yang2023doc} recursively expand and revise outlines, later work controls pacing, suspense, or actions~\citep{wang2023pacing,xie2024suspense,pei2024swag}, and multi-agent systems distribute planning, writing, and critique across roles~\citep{huot2025agentsroom,bae2024critics,chen2026storybox}. The outline, however, is fixed before any chapter is written, so details introduced during writing do not constrain later chapters.

\paragraph{Memory-based generation.}
Memory-based methods reuse earlier text through retrieval~\citep{zhou2023recurrentgpt}, compression~\citep{xia2025storywriter}, or full-text refinement~\citep{wu2026superwriter}, or maintain knowledge graphs of entities and events~\citep{wang2025dome,li2025storyteller}; general-purpose agent memories follow similar designs~\citep{park2023generative,packer2023memgpt,zhong2024memorybank,xu2025amem,rasmussen2025zep}. Closer to our work, FactTrack~\citep{lyu2025facttrack} tracks time-aware world facts, SCORE~\citep{yi2025score} and CHIRON~\citep{gurung2024chiron} maintain character and event representations, Octopus~\citep{wang2026octopus} binds persistent character and event memories to the writing context, and the Narrative World Model~\citep{saifullah2026narrativeworldmodel} organizes writer memory around narratological categories. DeepWriter~\citep{wang2026deepwriter} reaches book length, but it writes information-rich non-fiction grounded in retrieved external knowledge rather than novels. \NstAgent differs from these methods in the following ways: its state records prospective obligations alongside retrospective facts, the writer updates the state through a typed tool and can search fragments or read source chapters on demand, and we study how consistency changes as the same protocol scales from 10K to 100K words.

\paragraph{Evaluation.}
Story evaluation has moved from reference-based metrics and human ratings~\citep{guan2021openmeva,chhun2022hanna} toward LLM judges~\citep{zheng2023judging,liu2023geval,chhun2024enjoystories}. For writing quality, WritingBench~\citep{wu2025writingbench} scores responses against query-specific criteria, and recent benchmarks target book-length narratives~\citep{yang2025longstoryeval,wang2025novelbenchmark,fein2026litbench}. For consistency, ConStory-Bench~\citep{li2026constory} asks a judge to enumerate evidence-grounded contradictions in five categories and nineteen subtypes and reports their density per 10K words; related work detects plot holes~\citep{ahuja2025plotholes} and checks commitments in interactive narratives~\citep{ma2026ncpbench}. Because judges have limited recall over very long contexts~\citep{liu2024lostmiddle,chen2026longjudgebench}, Section~\ref{sec:eval-ext} extends ConStory-Bench so that consistency can be evaluated at longer lengths and remains comparable across lengths.

\section{Scaling Long-Form Story Generation}

\subsection{Problem Formulation}

\paragraph{Task.}
Given a story prompt \emph{p} and target length \emph{L}, the goal is to generate a long-form story \emph{S} by workflow $\pi_\theta$ with LLM $\theta$: 
\begin{equation}
S = \pi_\theta(p, L)
\end{equation}

\paragraph{Objective.}
The objective is to maximize the quality of the generated story \emph{S} with respect to the story prompt \emph{p} and target length \emph{L}: 
\begin{equation}
\max_{\pi_\theta}\mathbb{E}_{p, L}[Q(S)] \quad\text{with fixed}\quad\theta
\end{equation}

\paragraph{Quality and cost.}
We instantiate $Q$ with two complementary judgments, narrative consistency $Q_{\mathrm{con}}$ and writing quality $Q_{\mathrm{wq}}$, and do not collapse them into a single score. It is preferred to improve $\mathbb{E}_{p,L}[Q_{\mathrm{con}}(S)]$ and $\mathbb{E}_{p,L}[Q_{\mathrm{wq}}(S)]$ simultaneously, subject to
\begin{equation}
|S| \in [(1-\epsilon)L,\ (1+\epsilon)L],\qquad \mathrm{Cost}(\pi_\theta, L) = O(L),
\end{equation}
where $|S|$ is the word count and $\epsilon$ is the allowable relative error in word count. In this work, we set $\epsilon$ to 0.2 in all experiments. \emph{Length scaling} asks whether a single workflow $\pi_\theta$ can satisfy these requirements as $L$ grows without changing $\theta$.

\subsection{Scaling Challenges of Existing Methods}
\label{sec:challenge}

We examine three representative families (Table~\ref{tab:method-compare}) along the axes that become binding as $L$ grows: the cost of each generation step, whether information needed by later chapters survives, and whether the pipeline can reliably reach the target length.

\begin{table}[t]
\centering
\small
\setlength{\tabcolsep}{5pt}
\begin{tabular}{llllll}
\toprule
\textbf{Method} & \textbf{Memory Type} & \textbf{Update Type} & \textbf{Access} & \makecell[l]{\textbf{Future}\\\textbf{Obligations}} & \makecell[l]{\textbf{Context per}\\\textbf{Chapter}} \\
\midrule
Direct & -- & -- & -- & -- & Prompt only \\
DOME & Structured & Incremental & Passive & No & Outline; retrieved triples \\
StoryWriter & Free-text & Rewrite & Passive & No & Event plan; compressed history \\
RollSum & Free-text & Rewrite & Passive & Implicit & Outline; summary\\
\oursbg\NstAgent & \oursbg Structured & \oursbg Incremental & \oursbg Active & \oursbg Explicit & \oursbg Outline; state \\
\bottomrule
\end{tabular}
\caption{Memory design of the compared methods. \emph{Rewrite} regenerates the memory after each chapter, whereas \emph{incremental} updates add or replace individual entries. \emph{Passive} memory is supplied to the writer by the pipeline; \emph{active} memory lets the writer decide what to read or search. Direct writes the whole story in a single call.}
\label{tab:method-compare}
\end{table}

\paragraph{One-pass generation.}
Direct generation with long-output models must emit the entire story in a single call, and their main difficulty is simply reaching the requested length. Concretely, in our 10K experiments, only 21\% of first attempts from DeepSeek-V4-Flash and 74\% from GPT-5.6 Luna fall inside the $\pm 20\%$ acceptance band and an accepted story costs 1.9 and 1.3 calls on average. Scaling this approach further requires training longer-output models.

\paragraph{Knowledge-graph memory.}
Knowledge-graph methods such as DOME~\citep{wang2025dome} and STORYTELLER~\citep{li2025storyteller} extract entities and relations as chapters are written and retrieve them before the next one. This provides structure, but maintaining the memory is expensive: entity matching, graph queries, and per-triple LLM calls grow with the history, and even at 10K words DOME issues thousands of memory calls per story (Table~\ref{tab:efficiency}). The memory is also passive, since the pipeline decides what to retrieve by matching entities rather than by what the writer needs. We compare with DOME, whose released implementation fixes the story to five acts, so we evaluate it only at 10K.

\paragraph{Free-text memory.}
Rolling summaries and compressed histories~\citep{xia2025storywriter,chang2024booookscore} keep the context compact at a cost of roughly one extra call per chapter. The summary, however, is an unstructured string that is regenerated rather than edited. Each rewrite must decide which details to keep, details that seemed minor when written are compressed away before a later chapter needs them, and exact facts can silently change between versions (Appendix~\ref{app:case}). A structured state keeps the same compact footprint but changes how memory is maintained. Its entries are typed and keyed, character snapshots are overwritten rather than paraphrased, open obligations stay listed until they are resolved, and the writer can actively look up the source text when the state is not enough.

\subsection{Length-Comparable Consistency Evaluation}
\label{sec:eval-ext}
\paragraph{Consistency error density.}
ConStory-Bench~\citep{li2026constory} prompts an LLM judge with a story and one category at a time (characterization, factual detail, narrative style, timeline and plot, world building), and the judge returns contradictions with verbatim evidence under nineteen subtypes. Because longer stories offer more opportunities for error, ConStory-Bench does not score raw counts but the consistency error density (CED), the number of errors per 10K words, averaged over stories:
\begin{equation}
\mathrm{CED} = \frac{E}{W/10^4},
\end{equation}
where $E$ is the error count of a story and $W$ its word count; lower is better. Following its released code, we count $E_{\mathrm{sub}}$, the number of subtypes with at least one reported contradiction, and additionally report $E_{\mathrm{ins}}$, the total number of reported contradictions. We refer to the resulting metrics as Subtype CED and Instance CED.

\paragraph{Length normalization alone is not enough.}
ConStory-Bench targets stories of 8K--10K words, a range in which the number of reported errors grows roughly in proportion to story length~\citep{li2026constory}, so dividing by length removes the length bias. This assumption breaks down for much longer stories. When a judge reads an entire 50K- or 100K-word story, the number of contradictions it reports grows far more slowly than the story itself, because it cannot enumerate every contradiction in such a long text~\citep{liu2024lostmiddle,chen2026longjudgebench}. Full-story CED therefore falls as target length grows even when the writing does not improve: for an earlier version of our agent, it dropped by roughly 40\% from 20K to 50K words and again from 50K to 100K (Appendix~\ref{app:exp-details}). We instead keep the evaluated span close to the length for which ConStory-Bench was designed.

\paragraph{Fixed-size terminal windows.}
We insert a scope marker into the story and instruct the judge to report a contradiction only if its later manifestation lies after the marker, while using the entire preceding narrative as evidence. The marker is placed before the last $k\in\{10,7/8,5,4\}$ chapters for $L\in\{10\mathrm{K},20\mathrm{K},50\mathrm{K},100\mathrm{K}\}$. With 10, 15, 25, and 40 chapters at these lengths, the window covers roughly the final 10K words, so the judge enumerates errors over a span of similar size at every length, while each error is still checked against everything written before it. $W$ is the word count of the window, and the globally defined subtype is still checked over the full story (Appendix~\ref{app:protocol}). The modified judge instructions are given in Appendix~\ref{app:prompts}.

\section{\texorpdfstring{\NstAgent}{\NstAgent}}
\label{sec:method}

\begin{figure}[t]
    \centering
    \includegraphics[width=1.0\linewidth]{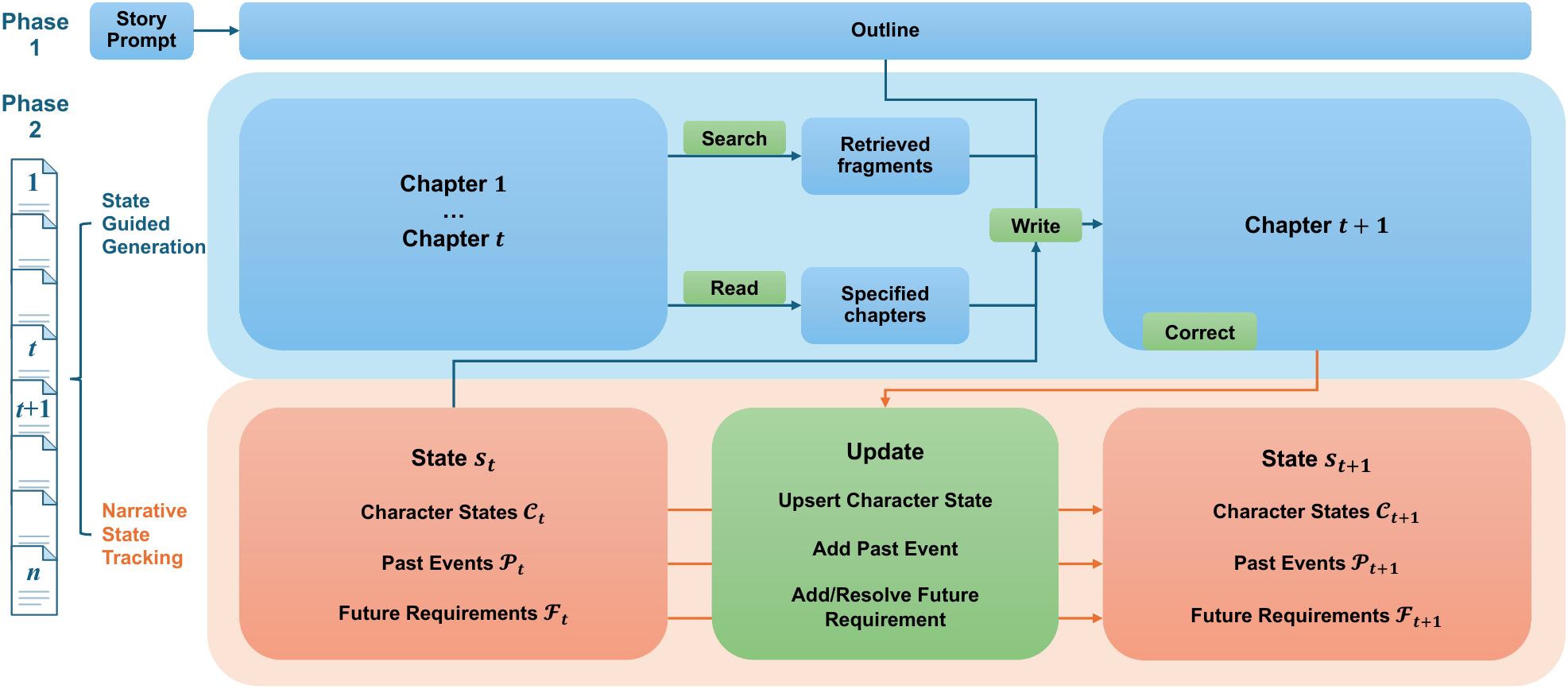}
    \caption{Overview of \NstAgent. Our framework first plans a frozen outline from the story prompt, and then writes the story chapter by chapter: in state-guided generation, the model writes chapter $t{+}1$ from the outline and state $t$, optionally searching or reading earlier chapters and correcting errors; in narrative state tracking, the update call turns state $t$ into state $t{+}1$.}
    \label{fig:overview}
\end{figure}

\subsection{Overview}

As shown in Figure~\ref{fig:overview}, \NstAgent writes a story in two phases. A planner first produces a premise and a chapter outline $O=(o_1,\dots,o_T)$, where each $o_t$ contains a title, a description, and a target word count $w_t$; the number of chapters is anchored to the target length. The outline is then frozen, and \NstAgent writes chapters sequentially while maintaining a narrative state $s_t$. For each chapter $t$, the model receives the original prompt $p$, the full outline $O$, and the current state $s_{t-1}$, writes the chapter $c_t$ with the help of tools, and finally submits a state update $\Delta_t$:
\begin{equation}
(c_t, \Delta_t) \sim \pi_\theta\big(\cdot \mid p,\ O,\ s_{t-1},\ t\big), \qquad s_t = \mathcal{U}(s_{t-1}, \Delta_t).
\end{equation}

The prompt for chapter $t$ contains no earlier chapter text. Earlier prose reaches the model through the state and, on demand, through read and search tools, so the per-chapter context depends on $|p|+|O|+|s_{t-1}|$ rather than on the length already written. Compared with a rolling summary, both see the same prompt and frozen outline, but \NstAgent carries typed, keyed entries that the model edits explicitly instead of a free-text summary that it rewrites, and it can additionally consult and correct earlier chapters.

Following the narratological views that a narrative is a sequence of state transitions~\citep{todorov1977poetics} and that it raises expectations it must later close~\citep{carroll2007closure}, the state is bidirectional: it records who the characters currently are and what has happened (retrospective), as well as what the story has promised but not yet delivered (prospective).

\subsection{State-Guided Generation}

\paragraph{Context.}
The chapter prompt presents, in order, the original story prompt, the full frozen outline, the current narrative state serialized as JSON, the number of completed chapters, and the current task: chapter id, title, description, and the target $w_t$ with the acceptance range. The prompt specifies an execution order: search or find earlier text if needed, write the chapter, correct inconsistencies if necessary, update the state once, and then end (Appendix~\ref{app:prompts}).

\paragraph{Tools.}
Generation is a tool-use loop~\citep{yao2023react} with four content tools.
\textbf{read}$(i)$ returns the full text of a written chapter $i$.
\textbf{search}$(q)$ splits the query into terms and returns at most a fixed number of sentence windows from earlier chapters ranked by BM25~\citep{robertson2009bm25}.
\textbf{write}$(t, \text{title}, \text{content})$ submits the complete chapter; the controller counts words and returns feedback whether the count is in the acceptance range, and a rejected draft does not consume the chapter's single successful write.
\textbf{correct}$(i, \text{old}, \text{new})$ replaces an exact span in the current or an earlier chapter to fix a factual or continuity error; it is limited per chapter and cannot be used for stylistic polishing. Read and search calls also have fixed budgets.

\paragraph{Length control.}
Length is enforced only through \textbf{write}. When a draft is rejected, the tool reports its actual word count, the target, and the accepted range, and the model rewrites the chapter. Because every chapter must pass this gate, the total length follows the outline without requiring any single call to produce more than one chapter. A response that stops at the output-token limit is never accepted, even if its visible text happens to fall within the range.

\subsection{Narrative State Tracking}

\paragraph{Narrative State.}
The narrative state $s_t=(\mathcal{C}_t,\mathcal{P}_t,\mathcal{F}_t)$ has three typed collections.
\emph{Character states} $\mathcal{C}_t$ map a character name to a snapshot of that character's current location, goal, relationships, knowledge, possessions, and physical or emotional condition~\citep{rashkin2018naivepsychology,gurung2024chiron}.
\emph{Past events} $\mathcal{P}_t$ are keyed records of completed events that are not explicit in the outline and may affect later plot.
\emph{Future requirements} $\mathcal{F}_t$ are keyed, unresolved obligations that later chapters must fulfill, such as a planted clue, an unanswered question, or a promised confrontation~\citep{xie2024suspense,carroll2007closure}.

\paragraph{Tracking.}
After the chapter is accepted, the model calls \textbf{update} exactly once with four native JSON arrays,
$\Delta_t = (\Delta^{\mathcal{C}}_t, \Delta^{\mathcal{P}}_t, \Delta^{\mathcal{F}+}_t, \Delta^{\mathcal{F}-}_t)$:
upsert\_character\_state (name, description); add\_past\_event (key, description); add\_future\_requirement (key, description); and resolve\_future\_requirement (keys). The transition $\mathcal{U}$ applies them atomically:
\begin{equation}
\mathcal{C}_t = \mathcal{C}_{t-1} \triangleleft \Delta^{\mathcal{C}}_t,\qquad
\mathcal{P}_t = \mathcal{P}_{t-1} \cup \Delta^{\mathcal{P}}_t,\qquad
\mathcal{F}_t = \big(\mathcal{F}_{t-1} \cup \Delta^{\mathcal{F}+}_t\big) \setminus \Delta^{\mathcal{F}-}_t,
\end{equation}
where $\triangleleft$ replaces the entry of an existing name and inserts a new one otherwise. Empty arrays are passed when a collection does not change.

\paragraph{Design choices.}
We encode the operation in the field name rather than asking the model to choose a collection, an operation, and a numeric id for every change. Characters are overwritten as complete snapshots, so they do not accumulate stale copies. Past events are filtered by the rule that the outline does not already state them, which keeps planned content out of the state. Requirements are resolved by stable keys the model itself chose. The detailed semantics of each field appear in the chapter prompt, while the tool schema only enforces structure.

\paragraph{State size.}
Character snapshots grow with the cast rather than the text, and requirements are removed once fulfilled. Past events accumulate but remain far shorter than the prose: even at 100K words, a final state holds about a dozen characters, fewer than ten open requirements, and past events amounting to a small fraction of the story (Appendix~\ref{app:more-results}). Since each chapter prompt contains the outline and the state but not earlier prose, and chapter length is fixed by the outline, the number of calls and generated tokens grow approximately linearly with the length of the novel.

\section{Experiments}

\subsection{Settings}

\paragraph{Data.}
We draw 100 English prompts from the generation task of ConStory-Bench~\citep{li2026constory}. Every prompt is written at target lengths of 10K, 20K, and 50K words; because judging 100K-word stories is costly, the 100K setting uses 50 prompts. DOME is evaluated on 20 prompts due to its high cost.

\paragraph{Backbones.}
We use two backbones from different model families, DeepSeek-V4-Flash and GPT-5.6 Luna, each in its default reasoning mode. Within a backbone, the same model plans, writes, and, for RollSum, summarizes. For every backbone, prompt, and target length, a premise and chapter outline are generated once and frozen, and all chapter-based methods read the same plan.

\paragraph{Baselines.}
At 10K words, we compare \NstAgent with four baselines covering the families in Section~\ref{sec:challenge}. (i) \textbf{Direct} generates the whole story in a single call. (ii) \textbf{DOME}~\citep{wang2025dome} writes chapters from a dynamic outline with knowledge-graph memory, and (iii) \textbf{StoryWriter}~\citep{xia2025storywriter} writes events with multiple agents and a compressed history; both follow their released pipelines. (iv) \textbf{RollSum} reads the same frozen outline as \NstAgent, writes each chapter as plain text, and rewrites a summary after every chapter, so the next chapter sees only that summary. Because the other baselines cannot reliably reach longer targets, the comparison from 10K to 100K words is between \NstAgent and RollSum. All methods share the same word-counting rule and a $\pm20\%$ length gate, applied per chapter for chapter-based methods and per story otherwise; configurations are listed in Appendix~\ref{app:exp-details}.

\paragraph{Evaluation.}
We measure consistency with the extended \textbf{ConStory-Bench} of Section~\ref{sec:eval-ext}, reporting Subtype CED and Instance CED, and writing quality with \textbf{WritingBench}~\citep{wu2025writingbench}, reporting the mean score over its five query-specific criteria on a 1--10 scale. Both benchmarks use DeepSeek-V4-Pro as the judge. A story is scored only if it is complete, all its chapters pass the length gate, and every judgment finishes; details are in Appendix~\ref{app:exp-details}. Appendix~\ref{app:protocol} verifies that the judge's recall does not decay over a 100K-word prefix.

\subsection{Results and Analysis}
\begin{table}[t]
\centering
\small
\resizebox{\linewidth}{!}{%
\begin{tabular}{llcccc}
\toprule
\textbf{Model} &\textbf{Method} &\textbf{Subtype CED ($\downarrow$)} & \textbf{Instance CED ($\downarrow$)} & \textbf{Writing Quality ($\uparrow$)} &\textbf{Avg. Words} \\
\midrule
\multirow{5}{*}{DeepSeek-V4-Flash} & Direct & 6.225 & 9.293 & 8.705 & 9,416 \\
& DOME & 9.607 & 16.765 & 4.870 & 9,990 \\
& StoryWriter & 7.395 & 11.803 & 7.353 & 9,830\\
& RollSum & \underline{5.981} & \underline{9.171} & \underline{8.946} & 10,186\\
& \oursbg\NstAgent & \oursbg\textbf{4.541} & \oursbg\textbf{6.392} & \oursbg\textbf{8.998} & \oursbg 11,365\\
\midrule
\multirow{5}{*}{GPT-5.6 Luna} & Direct & 5.428 & 7.961 & \underline{9.233} & 9,721 \\
& DOME & 6.861 & 11.419 & 6.950 & 10,080 \\
& StoryWriter & 5.163 & 7.679 & 8.408 & 10,304\\
& RollSum & \textbf{4.638} & \textbf{6.653} & 9.228 & 10,909\\
& \oursbg\NstAgent & \oursbg\underline{4.852} & \oursbg\underline{6.826} & \oursbg\textbf{9.258} & \oursbg 11,155\\
\bottomrule
\end{tabular}
}
\caption{Results on 10K. Best in \textbf{bold}, second best \underline{underlined}; average word counts are not ranked.}
\label{tab:10k}
\end{table}

\begin{table}[t]
\centering
\small
\resizebox{\linewidth}{!}{%
\begin{tabular}{llccccc}
\toprule
\textbf{Model} &\textbf{Method}  &\makecell[l]{\textbf{Target}\\\textbf{Length}} &\textbf{Subtype CED ($\downarrow$)} & \textbf{Instance CED ($\downarrow$)} & \textbf{Writing Quality ($\uparrow$)} &\textbf{Avg. Words} \\
\midrule
\multirow{8}{*}{DeepSeek-V4-Flash} & \multirow{4}{*}{RollSum} & 10K & 5.981 & 9.171 & 8.946 & 10,186 \\
&& 20K & 6.761 & 10.917 & 8.986 & 20,187 \\
&& 50K & 7.162 & 11.458 & 9.027 & 51,332 \\
&& 100K & 7.239 & 11.820 & 8.964 & 100,470 \\
& \oursbg & \oursbg 10K & \oursbg\textbf{4.541} & \oursbg\textbf{6.392} & \oursbg\textbf{8.998} & \oursbg 11,365 \\
& \oursbg & \oursbg 20K & \oursbg\textbf{5.382} & \oursbg\textbf{8.350} & \oursbg\textbf{9.140} & \oursbg 22,673 \\
& \oursbg & \oursbg 50K & \oursbg\textbf{5.793} & \oursbg\textbf{8.906} & \oursbg\textbf{9.240} & \oursbg 55,219 \\
& \oursbg\multirow{-4}{*}{\NstAgent} & \oursbg 100K & \oursbg\textbf{4.859} & \oursbg\textbf{7.239} & \oursbg\textbf{9.204} & \oursbg 108,295 \\
\midrule
\multirow{8}{*}{GPT-5.6 Luna} & \multirow{4}{*}{RollSum} & 10K & \textbf{4.638} & \textbf{6.653} & 9.228 & 10,909 \\
&& 20K & 6.057 & 8.972 & 9.166 & 20,573 \\
&& 50K & 6.014 & 9.095 & 9.196 & 50,696 \\
&& 100K & 5.765 & 9.369 & 9.064 & 100,773 \\
& \oursbg & \oursbg 10K & \oursbg 4.852 & \oursbg 6.826 & \oursbg\textbf{9.258} & \oursbg 11,155 \\
& \oursbg & \oursbg 20K & \oursbg\textbf{5.193} & \oursbg\textbf{7.698} & \oursbg\textbf{9.216} & \oursbg 22,076 \\
& \oursbg & \oursbg 50K & \oursbg\textbf{5.271} & \oursbg\textbf{7.695} & \oursbg\textbf{9.210} & \oursbg 48,748 \\
& \oursbg\multirow{-4}{*}{\NstAgent} & \oursbg 100K & \oursbg\textbf{4.887} & \oursbg\textbf{7.218} & \oursbg\textbf{9.244} & \oursbg 94,798 \\
\bottomrule
\end{tabular}
}
\caption{Results from 10K to 100K. Better of the two methods at each length in \textbf{bold}. 10K--50K use the 100 English prompts and 100K uses 50 of them.}
\label{tab:100k}
\end{table}

\paragraph{Performance on 10K.}
Table~\ref{tab:10k} compares all methods at 10K words, and Tables~\ref{tab:type-10k-ds} and~\ref{tab:type-10k-luna} break the results down by error category. On both backbones, \NstAgent and RollSum are more consistent than Direct, DOME, and StoryWriter, and Direct, RollSum, and \NstAgent write better than DOME and StoryWriter, with \NstAgent achieving the highest writing quality. On DeepSeek-V4-Flash, \NstAgent is the most consistent by a clear margin. On GPT-5.6 Luna, \NstAgent and RollSum are at parity: RollSum's instance CED is 2.5\% lower, but the difference is not significant (Appendix~\ref{app:stats}). We attribute this to the length: 10K words are short enough for a strong model to keep most of the narrative state in a free-text summary, so explicit tracking has little room to help. The next paragraph shows that this parity is specific to 10K.

\paragraph{Performance from 10K to 100K.}
Table~\ref{tab:100k} and Figure~\ref{fig:intro} extend the comparison with RollSum to 100K words, and Tables~\ref{tab:type-ds} and~\ref{tab:type-luna} break it down by error category. From 20K to 100K words, \NstAgent is more consistent and writes better than RollSum on both backbones, and its advantage becomes more pronounced as stories grow longer. On DeepSeek-V4-Flash, \NstAgent is more consistent than RollSum at every length, with the largest margin at 100K. On GPT-5.6 Luna, the two methods are comparable at 10K, but \NstAgent's advantage then grows steadily, lowering instance CED by 1.27, 1.40, and 2.15 at 20K, 50K, and 100K. Even a stronger backbone therefore struggles to maintain the narrative state in free-text memory as stories grow much longer, and explicit narrative state tracking substantially mitigates this.

Within \NstAgent, neither consistency nor writing quality degrades markedly as stories grow. Writing quality stays high on both backbones and on DeepSeek-V4-Flash even rises with length, possibly because longer stories develop richer plots. Error density also stays within a narrow range across lengths, rising only slightly beyond 10K and falling back at 100K. Part of this drop may come from judging 100K-word stories, although the injection study of Appendix~\ref{app:judge-recall} finds no loss of recall for explicit contradictions; either way, \NstAgent maintains the narrative state and its consistency over very long stories.

\paragraph{Generalization across backbones.}
DeepSeek-V4-Flash and GPT-5.6 Luna come from different providers and use the same agent loop very differently: DeepSeek-V4-Flash reasons at length and frequently reads and searches earlier chapters, whereas GPT-5.6 Luna produces several times fewer tokens, rarely searches, and relies mostly on the state it maintains (Appendix~\ref{app:more-results}). With the same prompts, tools, and state schema, the gains above hold on both. That they persist when one backbone barely uses retrieval, and that removing all lookback tools still leaves \NstAgent ahead of RollSum (Table~\ref{tab:ablation}), suggests that the benefit comes mainly from tracking the narrative state rather than from a particular tool-use strategy. The same loop also runs on a small open-weight model and can be further optimized with reinforcement learning (Appendix~\ref{app:rl}).

\begin{table}[t]
\centering
\begin{minipage}[t]{0.63\linewidth}
\vspace{0pt}
\centering
\small
\resizebox{\linewidth}{!}{%
\begin{tabular}{llrrrrr}
\toprule
\textbf{Method} & \textbf{Length} & \textbf{Calls} & \makecell[r]{\textbf{Input}\\\textbf{Tokens}} & \makecell[r]{\textbf{Cached}\\\textbf{Input}} & \makecell[r]{\textbf{Output}\\\textbf{Tokens}} & \makecell[r]{\textbf{Cost}\\\textbf{(USD)}} \\
\midrule
Direct & 10K & 2 & 8.6K & 0.0K & 32.5K & 0.02 \\
StoryWriter & 10K & 26 & 108.0K & 62.3K & 91.2K & 0.07 \\
DOME & 10K & 3,501 & 1,719.1K & 0.2K & 580.4K & 0.76 \\
\midrule
\multirow{4}{*}{RollSum} & 10K & 40 & 209.7K & 86.7K & 252.2K & 0.19 \\
& 20K & 56 & 386.5K & 162.6K & 473.0K & 0.36 \\
& 50K & 93 & 867.2K & 383.1K & 775.7K & 0.62 \\
& 100K & 171 & 2,314.5K & 1,162.2K & 1,710.6K & 1.39 \\
\midrule
\oursbg & \oursbg 10K & \oursbg 68 & \oursbg 711.7K & \oursbg 413.0K & \oursbg 159.1K & \oursbg 0.17 \\
\oursbg & \oursbg 20K & \oursbg 104 & \oursbg 1,466.1K & \oursbg 882.8K & \oursbg 273.0K & \oursbg 0.31 \\
\oursbg & \oursbg 50K & \oursbg 161 & \oursbg 3,174.1K & \oursbg 1,856.6K & \oursbg 507.4K & \oursbg 0.64 \\
\oursbg\multirow{-4}{*}{\NstAgent} & \oursbg 100K & \oursbg 259 & \oursbg 7,499.4K & \oursbg 4,495.2K & \oursbg 953.2K & \oursbg 1.32 \\
\bottomrule
\end{tabular}
}
\caption{Average generation cost per story on DeepSeek-V4-Flash, measured from the usage field of every API call on the first five prompts and averaged over completed stories. Output tokens include reasoning tokens, and cached input is the part of the input served from the provider's cache. Cost uses off-peak prices of USD 0.22, 0.007, and 0.66 per million uncached input, cached input, and output tokens.}
\label{tab:efficiency}
\end{minipage}\hfill
\begin{minipage}[t]{0.34\linewidth}
\vspace{0pt}
\centering
\includegraphics[width=\linewidth]{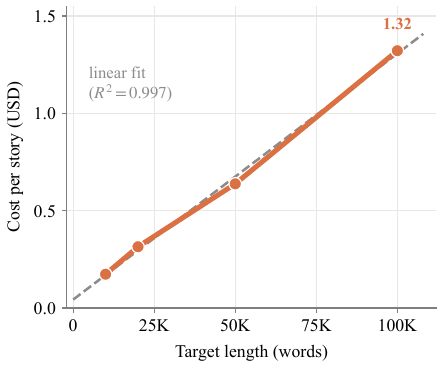}
\captionsetup{type=figure}
\caption{Cost of \NstAgent per story against target length, from the \NstAgent rows of Table~\ref{tab:efficiency}. Cost grows almost linearly with length, and the price per 10K words written stays between USD 0.13 and 0.17.}
\label{fig:cost}
\end{minipage}
\end{table}

\paragraph{Efficiency.}
Table~\ref{tab:efficiency} and Figure~\ref{fig:cost} report the cost of one story on DeepSeek-V4-Flash, measured from the usage field of every API call. The cost of \NstAgent grows approximately linearly with length: calls per chapter stay nearly constant, and the price per 10K words written stays between USD 0.13 and 0.17. RollSum costs about the same but spends its budget differently. \NstAgent sends far more input, because every turn resends the state, yet about 60\% of it is served from the provider's cache at roughly a thirtieth of the uncached price; RollSum's cost is dominated by output, because it regenerates a long summary, with reasoning, after every chapter. At 10K words, Direct is by far the cheapest but cannot reach longer targets, and DOME is the most expensive, taking about 3,500 calls per story to maintain its knowledge graph. Under prefix caching, tracking an explicit state therefore costs about as much as free-text memory at every length we test (Appendix~\ref{app:limitations}). Tool usage and state size are in Appendix~\ref{app:more-results}.

\begin{table}[t]
\centering
\small
\resizebox{\linewidth}{!}{%
\begin{tabular}{lcccc}
\toprule
\textbf{Variant} & \textbf{Subtype CED ($\downarrow$)} & \textbf{Instance CED ($\downarrow$)} & \textbf{Writing Quality ($\uparrow$)} & \makecell{\textbf{Reads and searches} \textbf{per story}} \\
\midrule
\oursbg\NstAgent & \oursbg\textbf{5.382} & \oursbg\textbf{8.350} & \oursbg\textbf{9.140} & \oursbg 31.4 \\
$-$State & 5.902 & 9.400 & 9.046 & 52.9 \\
$-$Lookback & 6.577 & 9.786 & 8.980 & -- \\
\midrule
RollSum (reference) & 6.761 & 10.917 & 8.986 & -- \\
\bottomrule
\end{tabular}}
\caption{Ablations on DeepSeek-V4-Flash at 20K words. $-$State removes the narrative state, its update tool, and the state-maintenance guidance; $-$Lookback removes the read, search, and correct tools. RollSum is shown for reference.}
\label{tab:ablation}
\end{table}

\paragraph{Ablation study.}

Table~\ref{tab:ablation} removes each of \NstAgent's two memory channels on DeepSeek-V4-Flash at 20K words, and Table~\ref{tab:type-ablation} breaks the results down by error category. Here, $-$State deletes the narrative state and its update tool, and $-$Lookback the read, search, and correct tools. Removing either channel makes stories significantly less consistent and lowers writing quality (Table~\ref{tab:ablation-stats}). Without the state, the writer reads and searches earlier chapters far more often, yet still makes more errors, most visibly in timeline and plot and in plot threads that are opened but never closed, which are exactly what the state records. Without lookback, errors rise further, particularly in factual details and timelines, which require checking the exact wording of earlier chapters rather than a summary of them. Both variants still outperform RollSum, and the full agent outperforms both, so the two channels are complementary.

\section{Conclusion and Future Work}

We framed long-form story generation as a length scaling problem and made narrative consistency comparable across lengths by measuring error density over fixed-size terminal windows. We proposed \NstAgent, a training-free agent that writes chapter by chapter while tracking a typed narrative state of characters, past events, and future requirements through tools. From 10K to 100K words and on two backbones, it reaches the target length, keeps writing quality and narrative consistency stable, and improves over free-text memory by a margin that grows with length. These results suggest that \NstAgent provides an effective approach to scaling story generation toward full-length novels. Building on the length-comparable evaluation and the narrative state tracking framework, future work can explore richer state representations, better revision mechanisms, and fine-tuning the backbone to track narrative state.

\subsection*{AI use statement}

In this work, we used generative AI tools to implement the proposed method and baselines in code and to
translate parts of the manuscript. We have not used generative AI tools to develop the conceptual
framework, propose or refine hypotheses, design or provide feedback on the methodology or experiments,
clean or reformat datasets, or interpret results, and synthetic data generation, mathematical claims,
proofs, and qualitative data analysis are not applicable to this work. Additionally, we used generative
AI tools to write and edit software code, identify relevant literature, and edit the manuscript to
improve its readability. We have reviewed all AI-assisted work: the authors revised all AI-assisted text
and translations, checked every suggested reference, and inspected and tested all AI-assisted code. We
take responsibility for the final content of this work, including text, claims, code, and artifacts
produced with the aid of generative AI.

\subsection*{Ethics statement}

This work involves no human subjects and no
personal data; the prompts come from the public ConStory-Bench generation task~\citep{li2026constory},
and all evaluated texts are fiction generated by language models. All scores come from LLM judges, which
can carry stylistic biases~\citep{zheng2023judging}, and the judge shares a model family with one
backbone, which may favor its outputs~\citep{panickssery2024selfpreference}; we therefore report
results on two backbones from different providers and discuss the judge's precision in
Appendix~\ref{app:limitations}. The generated stories are unfiltered and may inherit biases of their
backbones, so they should be reviewed before any publication. Cheaper book-length generation could also
be misused to mass-produce low-quality or deceptive text. We report the monetary cost of every setting
in Table~\ref{tab:efficiency}.

\subsection*{Reproducibility statement}

Section~\ref{sec:method} specifies the narrative state, tools, and update rule, and
Section~\ref{sec:eval-ext} the evaluation protocol. Appendix~\ref{app:exp-details} lists generation,
baseline, and judging settings, Appendix~\ref{app:prompts} gives all prompts, and
Appendix~\ref{app:stats} the statistical procedure. Our repository contains the code of
\NstAgent, the baselines and our patch to the released StoryWriter, both benchmarks, and the RL study
(Appendix~\ref{app:rl}); the frozen outlines and WritingBench criteria shared by all methods; the
generated stories; and the judge output of every evaluated
story. Its scripts recompute every table and figure from these files and check each case-study
quotation (Appendix~\ref{app:case}). Because backbones and judges are commercial APIs that may
change over time, individual stories and scores cannot be reproduced exactly.

\bibliography{iclr2027_conference}
\bibliographystyle{iclr2027_conference}
\newpage
\appendix

\section{Implementation Details}
\label{app:impl}

\subsection{Experiment Details}
\label{app:exp-details}

\paragraph{\NstAgent.}
The planner produces a premise of about 300 words and a chapter-level JSON outline with a title, a description, and a target word count for each chapter; for longer targets it additionally writes a detailed synopsis and an act-level outline in between. The recommended chapter count is interpolated between anchors of 10, 15, 25, and 40 chapters at 10K, 20K, 50K, and 100K words. Plans are cached per backbone, prompt id, and target length together with a SHA-256 digest, and all chapter-based methods read the cache in read-only mode. Each chapter is a tool-use conversation that ends when the model outputs \textbf{DONE} after a successful \textbf{write} and \textbf{update}; the loop is capped at 50 turns per chapter, and state and chapters are checkpointed after every chapter so that interrupted runs resume from the last completed chapter. Generation uses the backbone's default reasoning mode and temperature 0.7. The \textbf{update} tool enforces the four arrays of Section~\ref{sec:method} with a strict JSON schema that rejects additional properties.

Per chapter the writer may call \textbf{read} at most 3 times, \textbf{search} at most 5 times, and
\textbf{correct} at most 3 times; exceeding a budget returns an error that tells the model to proceed.
\textbf{search} indexes every written chapter as overlapping sentence windows (each window is one
sentence plus its immediate predecessor and successor, with the centre sentence weighted three times)
in an SQLite FTS5 index with its default BM25 parameters, and returns the 8 highest-ranked windows.
Chapters edited by \textbf{correct} are reindexed before the next search. Rejected drafts are not kept
in full: the first draft rejected by the length gate stays in the conversation as feedback, and a later
rejected draft replaces it only if it moves at least 10\% closer to the accepted range, otherwise its
text is dropped from the history and only its word count is retained. At most one rejected draft is
therefore in context at any time, which keeps a chapter with several retries from filling the
conversation with near-duplicate prose.

\paragraph{Baselines.}
All methods use the same word-counting rule (one word per contiguous ASCII word or per CJK character), check the finish reason, and reject any response whose completion tokens reach the requested limit. Chapter-based methods accept a chapter only within $\pm20\%$ of its target, and Direct and the adapted baselines accept a story only within $\pm20\%$ of the total target. \emph{Direct} performs one full-story call per attempt with a 32,768-token output limit, without memory or tools. \emph{RollSum} writes chapters as plain assistant text without tools, with a 32,768-token limit per chapter call. After each accepted chapter, it asks the model for an updated story-so-far summary given the previous summary and the new chapter; the prompt imposes no word limit, and each summary call has a 16,384-token output limit. \emph{DOME} keeps its dynamic hierarchical outline and temporal knowledge-graph retrieval; its fixed five-act structure typically yields few chapters. \emph{StoryWriter} keeps its event and sub-event decomposition with writer and critic agents (about five to ten events, three sub-events each, at most 50 rounds). For DOME and StoryWriter, every single-call output limit is 32,768 tokens, and our adaptations only add unified counting, integrity checks, failure feedback, resumption, and necessary bug fixes.

\paragraph{Evaluation.}
\emph{WritingBench.} DeepSeek-V4-Pro generates five query-specific criteria per prompt with the released criteria prompt~\citep{wu2025writingbench} (Appendix~\ref{app:prompts}) and scores the full story against each from 1 to 10; Writing Quality is the mean over the five criteria.

\emph{Extended ConStory-Bench.} The judge evaluates each story once per category, returning contradiction instances with an exact quote, a contradicting passage, and a subtype label from Table~\ref{tab:taxonomy}. We insert the scope marker of Appendix~\ref{app:prompts} before the first chapter at 10K (so every chapter is in scope) and before the last 7, 5, and 4 chapters at 20K, 50K, and 100K. The one exception is RollSum on DeepSeek-V4-Flash at 20K. Its chapters are shorter, so its last 7 chapters would span only 9.3K words against 10.5K for \NstAgent, and because the number of errors a judge reports saturates with the size of the window, a smaller window inflates CED. We therefore evaluate its last 8 chapters, which span 10.7K words and match \NstAgent's checked-word budget; reporting the smaller window would have favored \NstAgent. On GPT-5.6 Luna the gap is smaller (9.6K against 10.4K words), and we keep the last 7 chapters for both methods. The judge is DeepSeek-V4-Pro with streaming output, a 65,536-token output limit, and a 3,600-second request limit. Partial or unparsable judgments are not scored; a judgment interrupted by a sporadic service failure may be resumed once for the missing categories.

\emph{Why a fixed-size window.} Before adopting terminal windows, we judged complete stories written by an earlier version of our agent with DeepSeek-V4-Flash on 20 prompts. Over the full story, Subtype CED was 1.47, 0.87, and 0.50 at 20K, 50K, and 100K words, and Instance CED was 1.78, 1.32, and 0.91. Converting back to counts, the judge reported about 3.1, 4.5, and 5.1 error subtypes and 3.7, 6.8, and 9.3 contradictions per story, so a fivefold increase in length raised the reported count by less than a factor of three.

\emph{Why every chapter is marked at 10K.} The scope-restricted template tells the judge to report only evidence after the marker, so running it without a marker leaves the judge with an unsatisfiable instruction. On 20 \NstAgent stories at 10K from DeepSeek-V4-Flash, we manually adjudicated all alerts from two runs with the same judge, one without a marker and one with every chapter marked. Marking every chapter nearly tripled the number of alerts while leaving root-cause precision essentially unchanged, and it raised the coverage of confirmed root causes (relative to the pool found by either run) from about 38\% to about 91\%.

\begin{table}[t]
\centering
\small
\begin{tabular}{lp{0.68\linewidth}}
\toprule
\textbf{Category} & \textbf{Subtypes} \\
\midrule
Characterization & memory, knowledge, skill/power fluctuation, forgotten ability \\
Factual detail & appearance, nomenclature, quantitative \\
Narrative style & perspective, tone, style shift \\
Timeline and plot & absolute time, duration, simultaneity, causeless effect, causal logic, abandoned plot element \\
World building & core rules, social norms, geography \\
\bottomrule
\end{tabular}
\caption{The 19 error subtypes of ConStory-Bench grouped by category.}
\label{tab:taxonomy}
\end{table}

\begin{table}[t]
\centering
\small
\resizebox{\linewidth}{!}{%
\begin{tabular}{llcccccccc}
\toprule
\multirow{2}{*}{\textbf{Model}} & \multirow{2}{*}{\textbf{Length}} & \multicolumn{5}{c}{\textbf{Tool calls per story}} & \multicolumn{3}{c}{\textbf{Final state size}} \\
\cmidrule(lr){3-7}\cmidrule(lr){8-10}
 & & \textbf{read} & \textbf{search} & \textbf{write} & \textbf{correct} & \textbf{update} & \textbf{Characters} & \textbf{Past events} & \textbf{Open req.} \\
\midrule
\multirow{4}{*}{DeepSeek-V4-Flash} & 10K & 15.9 & 1.7 & 18.6 & 0.3 & 13.3 & 5.2 & 34.3 & 0.8 \\
& 20K & 25.1 & 6.3 & 33.0 & 0.4 & 20.6 & 5.7 & 50.0 & 1.7 \\
& 50K & 47.0 & 27.2 & 56.3 & 1.3 & 31.1 & 9.6 & 87.6 & 6.7 \\
& 100K & 61.5 & 17.3 & 81.3 & 2.3 & 54.3 & 11.8 & 158.0 & 9.2 \\
\midrule
\multirow{4}{*}{GPT-5.6 Luna} & 10K & 3.6 & 0.0 & 15.8 & 0.1 & 10.4 & 6.3 & 35.4 & 1.2 \\
& 20K & 7.1 & 0.1 & 19.7 & 0.1 & 15.4 & 6.9 & 49.8 & 3.6 \\
& 50K & 11.8 & 0.4 & 26.2 & 0.2 & 25.6 & 9.1 & 81.5 & 8.9 \\
& 100K & 28.4 & 1.7 & 43.5 & 0.1 & 41.7 & 11.3 & 123.4 & 8.6 \\
\bottomrule
\end{tabular}
}
\caption{Average \NstAgent tool calls per story (including rejected write and update attempts) and final state size. Stories have 10, 15, 25, and 40 chapters at 10K, 20K, 50K, and 100K words.}
\label{tab:tools}
\end{table}

\paragraph{Tool usage and state size.}
\label{app:more-results}
Table~\ref{tab:tools} reports the average tool usage and final state size of \NstAgent; generation cost is in Table~\ref{tab:efficiency}. These counts come from the chapter loop of the 100-prompt runs rather than from API usage logs, so they are not directly comparable with that table. GPT-5.6 Luna generates several times fewer tokens than DeepSeek-V4-Flash. On Luna, \NstAgent uses 39.9, 57.3, 89.0, and 154.4 calls and 38.1K, 61.6K, 106.8K, and 204.0K output tokens per story at 10K, 20K, 50K, and 100K words (from 3.4K down to 2.2K tokens per 1K words), while RollSum's chapter-writing calls alone produce at least 43.3K, 70.2K, 133.8K, and 287.1K output tokens. Write calls exceed the number of chapters because drafts rejected by the length gate are rewritten. Both backbones rarely use \textbf{correct}, so consistency gains come mainly from state-guided writing rather than post-hoc repair. GPT-5.6 Luna almost never searches, whereas DeepSeek-V4-Flash searches increasingly at longer lengths. Past events grow by about four entries per chapter and are the main source of state growth.

\subsection{Prompt Templates}
\label{app:prompts}

\paragraph{\NstAgent prompts.}
The system prompt and the user prompt for each chapter are shown below. Placeholders are written in \texttt{\{braces\}}.

\begin{promptbox}{\NstAgent system prompt}
You are an experienced novelist writing a long novel chapter by chapter. Use the provided tools to read written chapters, search for specified content, write the current chapter, fix inconsistencies, and maintain the structured narrative state.
\end{promptbox}

\begin{promptbox}{\NstAgent chapter prompt (user message)}
You are writing Chapter \texttt{\{t\}} of a long novel.

\textbf{\# Original Story Prompt}\\
\texttt{\{prompt\}}

\textbf{\# Full Frozen Outline}\\
\texttt{\{outline JSON\}}

\textbf{\# Current Narrative State}\\
\texttt{\{state JSON\}}

\textbf{\# Completed Chapters}\\
The first \texttt{\{n\}} chapters have been completed.

\textbf{\# Current Task}\\
Chapter ID: \texttt{\{t\}}\\
Chapter Name: \texttt{\{title\}}\\
Description: \texttt{\{description\}}\\
Target word count for this chapter: \texttt{\{w\_t\}}. Keep the chapter strictly within $\pm$20\% of this target word count.

\textbf{Execution order:}\\
1.~Use \textbf{read} to revisit a specific chapter, or \textbf{search} to locate prior facts, as needed.\\
2.~Use \textbf{write} to write the complete current chapter.\\
3.~If necessary, use \textbf{correct} to precisely fix consistency errors in the current or a prior chapter.\\
4.~Once the prose is final, make exactly one \textbf{update} call containing character-state upserts, newly established completed events, new future requirements, and resolved future-requirement keys.\\
5.~When everything is complete, make no tool call and output only \textbf{DONE}; any other text does not finish the chapter.

\textbf{Writing and state requirements:}\\
-- The outline is a plan, not prose; expand it into scenes with dialogue, sensory detail, and internal experience.\\
-- In upsert\_character\_state, provide \texttt{\{name, description\}} with each character's complete current location, goal, relationships, knowledge, possessions, and physical or emotional condition; an existing name is replaced rather than appended. In add\_past\_event, record a completed event only when the fact is not already explicit in the frozen outline and may affect later plot; give it a stable snake\_case key. Add only concrete later obligations to add\_future\_requirement, also with stable keys, and pass the keys of requirements fulfilled in this chapter to resolve\_future\_requirement. Every field must be a native JSON array; never serialize or quote an array as a string. Include all four arrays and use \texttt{[]} when one has no changes.\\
-- Follow each tool's own conditions, limits, and schema.
\end{promptbox}

\paragraph{Planner prompts.}
The planner runs in up to four stages; the first and the last are used at every length, and the
synopsis and act stages are inserted for targets above 10K words. Each stage receives the original
story prompt and the output of the previous stage.

\begin{promptbox}{Planner, stage 1: premise}
Story prompt: \texttt{\{prompt\}}

The final novel will be about \texttt{\{L\}} words. Write a high-level premise of approximately 300
words. Cover the core conflict, protagonist, setting, central theme, and the overall narrative arc
from beginning to end. Do not break it into chapters yet.

Output ONLY the premise text, no headings or extra commentary.
\end{promptbox}

\begin{promptbox}{Planner, stage 2: synopsis (targets above 10K words)}
Story prompt: \texttt{\{prompt\}} \quad High-level premise: \texttt{\{premise\}}

Expand the premise into a detailed synopsis of approximately \texttt{\{n\}} words. The synopsis should
walk through the major story beats in order, introduce the principal characters and their motivations,
describe key locations, and lay out the major turning points and resolution. Do not break it into
chapters yet.

Output ONLY the synopsis text.
\end{promptbox}

\begin{promptbox}{Planner, stage 3: acts (targets above 10K words)}
Story prompt: \texttt{\{prompt\}} \quad Premise: \texttt{\{premise\}} \quad Synopsis: \texttt{\{synopsis\}}

This is a novel of about \texttt{\{L\}} words. Divide the story into a small number of acts or volumes
(typically 3--6). Give each act a title and a description of two to four sentences summarizing what
happens in it. The acts together must cover the entire synopsis with no gaps.

Output the act-level outline as readable text, with each act clearly numbered and titled.
\end{promptbox}

The final stage turns the act outline into the chapter-level JSON outline used by all chapter-based
methods: for each chapter an id, a title, a description, and a target word count, with the chapter
count anchored as described above.

\paragraph{RollSum summarizer prompt.}
After each accepted chapter, RollSum replaces its summary with the output of this call. The summary is
method-internal state and never counts toward the story's word count.

\begin{promptbox}{RollSum summarizer (system, then user)}
\textbf{System.} Maintain a compact rolling summary of a long novel. Preserve the character
situations, key events, relationship changes, timeline, and unresolved threads needed for later
chapters. Compress settled earlier information instead of retelling scenes, dialogue, or prose.

\textbf{User.} Original story prompt: \texttt{\{prompt\}}

Previous story-so-far summary: \texttt{\{summary\}}

Newly completed Chapter \texttt{\{t\}}, \texttt{\{title\}}: \texttt{\{chapter text\}}

Output only the updated story-so-far summary directly. Do not critique the chapter, offer writing
advice, or add other meta-text. Choose the summary length needed to preserve all information that may
matter to later chapters, while remaining concise and never continuing the story prose.
\end{promptbox}

\paragraph{WritingBench prompts.}
Writing quality uses the released WritingBench pipeline~\citep{wu2025writingbench} unchanged. The judge
first turns a prompt into five query-specific criteria, each with a description and five scoring bands,
and then scores a story against each criterion. Criteria are generated once per prompt and cached, so
every method, backbone, and target length is scored against identical rubrics; regenerating them per
method would let a method be scored against an easier rubric.

\begin{promptbox}{WritingBench criteria-generation prompt (verbatim)}
\emph{System.} You are an expert evaluator with extensive experience in evaluating the response of a
given query.

\emph{User.} Please generate five strict evaluation criteria for assessing the response given the
following query. Each criterion should include the following fields: name, criteria\_description, 1-2,
3-4, 5-6, 7-8, 9-10. The criteria should be designed to emphasize detailed assessment and distinguish
subtle differences in quality. Ensure that the criteria can discern issues such as relevance,
coherence, depth, specificity, and adherence to the query context. Do not include any additional text.
Only output the criteria in the specified JSON format.

** Query ** \texttt{\{query\}}

** Output format ** \texttt{[\{"name": "first\_criteria\_name", "criteria\_description": "\dots",
"1-2": "\dots", \dots, "9-10": "\dots"\}, \dots]}
\end{promptbox}

\begin{promptbox}{WritingBench scoring prompt (verbatim, abridged)}
\emph{System.} You are an expert evaluator with extensive experience in evaluating response of given
query.

\emph{User.} Evaluate the Response based on the Query and Criteria provided following the Scoring
Rules.

** Scoring Rules ** ``1-2'': critical deficiencies and major issues that prevent adequate
functionality. \dots\ ``9-10'': exceptional performance with all aspects optimally addressed.

-- Provide reasons for each score by indicating specific strengths or deficiencies within the Response.
Reference exact text passages to justify the score \dots

-- Be very STRICT and do not be misled by format or length; ensure that the Response is thoroughly
evaluated beyond superficial appearances.

-- Carefully discern whether the content of the Response is an illusion, appearing substantial but
actually entirely fabricated.

-- Sometimes the model may only provide an introduction or an overview without truly completing the
query, which should be considered a failed response. \dots\ Scoring Range: assign an integer score
between 1 and 10.

** Criteria ** \texttt{\{criteria\}} \quad ** Query ** \texttt{\{query\}} \quad ** Response **
\texttt{\{response\}}

Provide your evaluation based on the criteria restated below: \texttt{\{criteria\}}

Return the results in JSON: \texttt{\{"score": integer, "reason": "\dots"\}}
\end{promptbox}

The criteria are prompt-specific, so they differ in what they reward. The box below shows two of the
five generated for one ConStory-Bench prompt, abridged to the outer bands.

\begin{promptbox}{Example of generated criteria (abridged)}
\textbf{Dialogue Purity and Format Compliance.} Evaluates strict adherence to the ``dialogue only, no
speaker tags or narration'' requirement. \emph{1--2.} Heavy reliance on narration or speaker tags;
dialogue is interspersed with descriptive prose or attributions, violating the core format
requirement. \emph{9--10.} Flawless dialogue execution: absolutely no speaker tags, narration, or
descriptive cues; the entire narrative is conveyed purely through exchanges.

\textbf{Revelation of Stranger's Nature and Purpose.} Assesses how effectively the dialogue slowly
reveals the stranger's true nature and purpose, measuring the gradual unfolding, subtlety, and logical
consistency of the revelations within the conversation. \dots
\end{promptbox}

\paragraph{Extended ConStory-Bench judge prompts.}
We keep the original ConStory-Bench templates and output schema and change three parts, shown below for the characterization template. The timeline-and-plot template additionally states that abandoned\_plot\_elements remains a check over the full narrative, with \texttt{exact\_quote} allowed anywhere.

\begin{promptbox}{(1) Scope marker inserted into the story before the first target chapter}
\texttt{>>>>>>>>> TARGET ENDING CHAPTERS START >>>>>>>>>}

IMPORTANT: The following chapters are the only target chapters for error attribution. Continue using the entire preceding narrative as reference evidence, but report an error only when its later contradictory passage appears in one of the following chapters. For every reported item except the Timeline check's global abandoned\_plot\_elements, \texttt{exact\_quote} must be copied verbatim from text after this marker. If only the \texttt{contradiction\_pair} is after this marker, swap the two evidence fields; if neither is after this marker, omit the item.

Target chapter IDs: \texttt{\{target\_chapter\_ids\}}

\texttt{>>>>>>>>>>>>>>>>>>>>>>>>>>>>>>>>>>>>>>>>>}
\end{promptbox}

\begin{promptbox}{(2) Task description and scope block added to each category template}
\textbf{\# TASK:} Given a novel, novella, screenplay, or other extended narrative, identify and extract ALL character consistency errors whose later contradictory manifestation is anchored in the target ending chapters.

\textbf{\# TERMINAL-WINDOW SCOPE:}\\
$\bullet$ The full narrative is reference evidence, but only the ending chapters marked inside the story are evaluation targets.\\
$\bullet$ Report an error only when its later contradictory manifestation occurs in a target ending chapter.\\
$\bullet$ Earlier conflicting evidence may appear anywhere before it in the full narrative.\\
$\bullet$ In every reported item, \texttt{exact\_quote} MUST be the later target-chapter passage; \texttt{contradiction\_pair} may come from earlier in the full narrative.\\
$\bullet$ If the target-chapter evidence is currently treated as \texttt{contradiction\_pair}, swap the two evidence fields.\\
$\bullet$ Before returning JSON, verify that every \texttt{exact\_quote} occurs verbatim after the target marker; otherwise omit the item.\\
$\bullet$ Do not report errors whose later manifestation lies outside the target ending chapters.\\
$\bullet$ Target ending chapter IDs: \texttt{\{\{ Target Chapter IDs \}\}}
\end{promptbox}

\begin{promptbox}{(3) Closing reminder replacing the original final instruction}
\textbf{\# FINAL SCOPE REMINDER}

Use the full narrative only to establish and verify earlier evidence. Return only errors whose later contradictory manifestation (\texttt{exact\_quote}) is located in one of the target ending chapters: \texttt{\{\{ Target Chapter IDs \}\}}. Before returning JSON, verify that each \texttt{exact\_quote} can be copied verbatim from text after the target marker. If only \texttt{contradiction\_pair} is after the marker, swap the two evidence fields. Do not assign a target-chapter location to a quote from before the marker; omit any item whose \texttt{exact\_quote} remains outside the target chapters. Return the original JSON schema only.
\end{promptbox}

\section{Validation of the Evaluation Protocol}
\label{app:protocol}

\subsection{Judge Recall over Long Prefixes}
\label{app:judge-recall}

Terminal windows bound what the judge must enumerate, but not what it must remember: at 100K words it
still has to find the contradicted fact somewhere in up to 90K words of preceding text. If its recall
decayed with prefix length, the stable CED we report at 100K could be an artifact of the judge missing
errors rather than of the stories containing fewer. Long-context benchmarks report exactly this kind of
degradation~\citep{bai2025longbenchv2,wu2025longgenbench}, and audits of evaluation protocols for
long-form writing find their reliability is often assumed rather than measured~\citep{mei2026illusions}.
We test it directly.

\paragraph{Setup.}
In real \NstAgent stories we inject a pair of explicit, self-contained statements that contradict each
other, one of the five categories per injection. The later half always lands \emph{inside} the
evaluation window, so the protocol is obliged to report it, and the earlier half is placed at one of
three prefix depths: immediately before the window, at the midpoint of the prefix, or in the opening
chapter. The two halves are therefore separated by between zero and about 101K words. Each injection
carries a unique marker that occurs exactly twice in its story, so detection is an exact string match
in the judge's output rather than a semantic judgment. Per length we inject 30 contradictions (five
categories $\times$ three depths $\times$ two stories) and 6 control pairs whose two statements are
mutually consistent, and judge all 144 stories with the protocol and judge of the main experiments.

\paragraph{Results.}
Recall does not decay (Table~\ref{tab:recall}). The targeted category is reported for 119 of 120
injected contradictions, the single miss being at 10K, and every one of the 120 is reported under some
category, so all twelve length $\times$ depth cells are at 10/10 for detection, including the cell where
the two halves are about 101K words apart. Neither the target length nor the distance between the
halves predicts detection (Kendall $\tau=+0.11$, $p=0.18$, and $\tau=+0.06$, $p=0.46$). An earlier calibration with a weaker judge behaved very differently, falling
from 93\% at 10K--20K to 53\% at 50K--100K, so this is a property of the judge we use, not of the task.

Two caveats limit what this establishes. The injected statements are explicit and conspicuous, and
recall is at ceiling, so the experiment rules out the failure mode it tests but says nothing about
subtle, naturally occurring contradictions. And the control pairs expose a precision problem: 25\% of
them are reported as a contradiction of the targeted type, and 79\% are reported under some subtype,
mostly as a style shift caused by the inserted text itself. In the same runs, one injected
contradiction is filed under about four subtypes on average, against 1.5 for controls, which is direct
evidence that instance CED counts one underlying error several times and that subtype CED is the more
conservative measure.

\begin{table}[h]
\centering
\small
\begin{tabular}{lcccc}
\toprule
\textbf{Target length} & \makecell{\textbf{Prefix separating}\\\textbf{the two halves}} & \makecell{\textbf{Target subtype}\\\textbf{recall}} & \makecell{\textbf{Target category}\\\textbf{recall}} & \makecell{\textbf{Controls reported}\\\textbf{as target type}} \\
\midrule
10K & 0--9.3K & 28/30 & 29/30 & 0/6 \\
20K & 4.4K--15.3K & 30/30 & 30/30 & 2/6 \\
50K & 4.5K--46.8K & 29/30 & 30/30 & 2/6 \\
100K & 5.4K--101.0K & 30/30 & 30/30 & 2/6 \\
\bottomrule
\end{tabular}
\caption{Detection of injected contradictions by DeepSeek-V4-Pro under the terminal-window protocol.
The second column gives the range of mean word distances between the two halves across the three
injection depths. Controls contain no contradiction, so any report is a false positive.}
\label{tab:recall}
\end{table}

\subsection{Window-Local and Global Errors}
\label{app:global-local}

Eighteen of the nineteen subtypes are reported only if the contradiction manifests inside the terminal
window, but abandoned\_plot\_elements is checked over the whole story, so at 100K words it can
accumulate over 40 chapters while being divided by the roughly 10K words of the window. This inflates
its weight as the target length grows, and it does: the share of instances contributed by that subtype
rises from 5.1\% to 11.8\% on DeepSeek-V4-Flash and from 15.0\% to 26.4\% on GPT-5.6 Luna between 10K
and 100K words (Table~\ref{tab:global-local}). Reporting only the eighteen window-local subtypes lowers
every number but leaves the comparison intact (Table~\ref{tab:stats-scaling}, fourth column). We therefore keep the released definition in the main
tables and report the decomposition here rather than renormalizing one subtype by a different
denominator.

\begin{table}[h]
\centering
\small
\begin{tabular}{llcccc}
\toprule
\textbf{Model} & \textbf{Method} & \textbf{10K} & \textbf{20K} & \textbf{50K} & \textbf{100K} \\
\midrule
\multirow{2}{*}{DeepSeek-V4-Flash} & \NstAgent & 6.06 / 0.33 & 7.82 / 0.53 & 8.14 / 0.77 & 6.39 / 0.85 \\
 & RollSum & 8.49 / 0.68 & 10.21 / 0.70 & 10.25 / 1.21 & 10.64 / 1.18 \\
\midrule
\multirow{2}{*}{GPT-5.6 Luna} & \NstAgent & 5.80 / 1.02 & 6.64 / 1.06 & 6.15 / 1.55 & 5.31 / 1.90 \\
 & RollSum & 5.79 / 0.86 & 7.78 / 1.19 & 7.43 / 1.67 & 7.43 / 1.94 \\
\bottomrule
\end{tabular}
\caption{Instance CED split into window-local subtypes and the globally scoped
abandoned\_plot\_elements subtype (local / global). The two add up to the instance CED of
Table~\ref{tab:100k}.}
\label{tab:global-local}
\end{table}

\section{Additional Results}
\label{app:extra}

\subsection{Errors by Type}
\label{app:category}

Tables~\ref{tab:type-10k-ds}--\ref{tab:type-luna} report instance CED for all five error categories
and all nineteen subtypes, first at 10K words for all five methods and then from 20K to 100K words for
\NstAgent and RollSum. At 10K words, \NstAgent is the lowest of the five methods in every category and in
15 of the 19 subtypes on DeepSeek-V4-Flash; on GPT-5.6 Luna, where the aggregate is at parity with
RollSum, it is lowest in factual detail and below Direct, DOME, and StoryWriter in four of the five
categories. Beyond 10K words, \NstAgent has the lower error density in all five categories in all six
settings (30 of 30 comparisons) and in 13 to 17 of the 19 subtypes. The largest reductions sit in subtypes that require recalling an exact earlier fact: on
DeepSeek-V4-Flash at 100K, memory contradictions fall from 2.09 to 1.05, appearance mismatches from 1.32
to 0.65, and nomenclature confusions from 0.75 to 0.19, which are the facts a character snapshot stores
verbatim and a rewritten summary tends to paraphrase (Appendix~\ref{app:case}). In the ablations (Table~\ref{tab:type-ablation}), removing the state mainly raises timeline and plot errors, including unclosed plot threads, and removing lookback raises timeline and factual-detail errors.

\begin{table}[p]
\centering
\small
\resizebox{0.85\linewidth}{!}{%
\begin{tabular}{llccccc}
\toprule
\textbf{Category} & \textbf{Subtype} & \textbf{Direct} & \textbf{DOME} & \textbf{StoryWriter} & \textbf{RollSum} & \textbf{\NstAgent} \\
\midrule
\multirow{5}{*}{Characterization} & memory & 1.10 & 2.17 & 1.05 & 1.18 & \textbf{0.64} \\
 & knowledge & 0.09 & 0.10 & 0.13 & 0.09 & \textbf{0.06} \\
 & skill/power & 0.12 & 0.31 & 0.11 & 0.06 & \textbf{0.05} \\
 & forgotten ability & \textbf{0.00} & \textbf{0.00} & 0.01 & \textbf{0.00} & 0.01 \\
 & \cellcolor{nstbg}\textit{all} & \cellcolor{nstbg}1.30 & \cellcolor{nstbg}2.58 & \cellcolor{nstbg}1.30 & \cellcolor{nstbg}1.33 & \cellcolor{nstbg}\textbf{0.75} \\
\midrule
\multirow{4}{*}{Factual detail} & appearance & 0.73 & 1.12 & 1.00 & 0.89 & \textbf{0.62} \\
 & nomenclature & 0.47 & 1.57 & 0.64 & 0.39 & \textbf{0.28} \\
 & quantitative & 1.56 & \textbf{1.02} & 1.29 & 1.56 & 1.26 \\
 & \cellcolor{nstbg}\textit{all} & \cellcolor{nstbg}2.76 & \cellcolor{nstbg}3.71 & \cellcolor{nstbg}2.93 & \cellcolor{nstbg}2.85 & \cellcolor{nstbg}\textbf{2.16} \\
\midrule
\multirow{4}{*}{Narrative style} & perspective & 0.13 & 0.35 & 0.18 & 0.13 & \textbf{0.03} \\
 & tone & 0.04 & 0.05 & 0.06 & \textbf{0.01} & \textbf{0.01} \\
 & style shift & 0.25 & 1.67 & 1.05 & 0.38 & \textbf{0.04} \\
 & \cellcolor{nstbg}\textit{all} & \cellcolor{nstbg}0.42 & \cellcolor{nstbg}2.08 & \cellcolor{nstbg}1.29 & \cellcolor{nstbg}0.52 & \cellcolor{nstbg}\textbf{0.08} \\
\midrule
\multirow{7}{*}{Timeline and plot} & absolute time & 0.31 & 0.25 & 0.24 & 0.25 & \textbf{0.21} \\
 & duration & 0.99 & 1.02 & 0.77 & 0.91 & \textbf{0.71} \\
 & simultaneity & 0.23 & 0.31 & 0.28 & 0.23 & \textbf{0.12} \\
 & causeless effect & 0.26 & 0.46 & 0.59 & \textbf{0.20} & 0.22 \\
 & causal logic & 0.88 & 2.08 & 1.26 & 0.71 & \textbf{0.53} \\
 & abandoned plot & 0.63 & 1.98 & 1.38 & 0.68 & \textbf{0.33} \\
 & \cellcolor{nstbg}\textit{all} & \cellcolor{nstbg}3.30 & \cellcolor{nstbg}6.10 & \cellcolor{nstbg}4.54 & \cellcolor{nstbg}2.98 & \cellcolor{nstbg}\textbf{2.10} \\
\midrule
\multirow{4}{*}{World building} & core rules & 1.01 & 1.22 & 1.07 & 0.90 & \textbf{0.83} \\
 & social norms & 0.19 & 0.35 & 0.23 & \textbf{0.18} & \textbf{0.18} \\
 & geography & 0.31 & 0.71 & 0.44 & 0.41 & \textbf{0.29} \\
 & \cellcolor{nstbg}\textit{all} & \cellcolor{nstbg}1.51 & \cellcolor{nstbg}2.29 & \cellcolor{nstbg}1.74 & \cellcolor{nstbg}1.49 & \cellcolor{nstbg}\textbf{1.30} \\
\midrule
\textbf{Total} & & 9.29 & 16.76 & 11.80 & 9.17 & \textbf{6.39} \\
\bottomrule
\end{tabular}}
\caption{Instance CED by error category and subtype at 10K words on DeepSeek-V4-Flash, in errors per 10K checked words. Rows marked \textit{all} (shaded) give the category total. The lowest value of each row is in \textbf{bold}; totals equal the instance CED of Table~\ref{tab:10k}. \NstAgent is lowest in all five categories and in 15 of the 19 subtypes, counted on unrounded values.}
\label{tab:type-10k-ds}
\end{table}

\begin{table}[p]
\centering
\small
\resizebox{0.85\linewidth}{!}{%
\begin{tabular}{llccccc}
\toprule
\textbf{Category} & \textbf{Subtype} & \textbf{Direct} & \textbf{DOME} & \textbf{StoryWriter} & \textbf{RollSum} & \textbf{\NstAgent} \\
\midrule
\multirow{5}{*}{Characterization} & memory & 0.82 & 1.65 & 0.66 & \textbf{0.53} & 0.68 \\
 & knowledge & 0.01 & 0.05 & 0.01 & 0.03 & \textbf{0.00} \\
 & skill/power & 0.09 & 0.10 & 0.12 & 0.08 & \textbf{0.07} \\
 & forgotten ability & 0.00 & 0.00 & 0.00 & 0.00 & 0.00 \\
 & \cellcolor{nstbg}\textit{all} & \cellcolor{nstbg}0.92 & \cellcolor{nstbg}1.80 & \cellcolor{nstbg}0.79 & \cellcolor{nstbg}\textbf{0.64} & \cellcolor{nstbg}0.76 \\
\midrule
\multirow{4}{*}{Factual detail} & appearance & \textbf{0.57} & 0.95 & 0.97 & 0.93 & 0.69 \\
 & nomenclature & \textbf{0.40} & 1.20 & 0.59 & 0.42 & 0.59 \\
 & quantitative & 1.55 & 0.95 & \textbf{0.70} & 0.99 & 0.82 \\
 & \cellcolor{nstbg}\textit{all} & \cellcolor{nstbg}2.52 & \cellcolor{nstbg}3.10 & \cellcolor{nstbg}2.26 & \cellcolor{nstbg}2.34 & \cellcolor{nstbg}\textbf{2.11} \\
\midrule
\multirow{4}{*}{Narrative style} & perspective & \textbf{0.01} & 0.50 & 0.06 & \textbf{0.01} & 0.08 \\
 & tone & \textbf{0.00} & \textbf{0.00} & 0.01 & \textbf{0.00} & 0.01 \\
 & style shift & \textbf{0.02} & 0.75 & 0.49 & 0.17 & 0.17 \\
 & \cellcolor{nstbg}\textit{all} & \cellcolor{nstbg}\textbf{0.03} & \cellcolor{nstbg}1.25 & \cellcolor{nstbg}0.56 & \cellcolor{nstbg}0.18 & \cellcolor{nstbg}0.26 \\
\midrule
\multirow{7}{*}{Timeline and plot} & absolute time & 0.28 & 0.25 & \textbf{0.17} & 0.20 & 0.24 \\
 & duration & 0.89 & 0.85 & 0.47 & 0.53 & \textbf{0.44} \\
 & simultaneity & 0.31 & 0.35 & 0.23 & \textbf{0.13} & 0.16 \\
 & causeless effect & \textbf{0.26} & 0.30 & 0.31 & 0.33 & 0.33 \\
 & causal logic & 0.66 & 1.20 & 0.73 & \textbf{0.48} & 0.49 \\
 & abandoned plot & \textbf{0.71} & 1.15 & 1.09 & 0.86 & 1.02 \\
 & \cellcolor{nstbg}\textit{all} & \cellcolor{nstbg}3.10 & \cellcolor{nstbg}4.10 & \cellcolor{nstbg}3.00 & \cellcolor{nstbg}\textbf{2.52} & \cellcolor{nstbg}2.67 \\
\midrule
\multirow{4}{*}{World building} & core rules & 0.90 & \textbf{0.45} & 0.62 & 0.63 & 0.62 \\
 & social norms & 0.17 & 0.35 & 0.23 & \textbf{0.13} & 0.21 \\
 & geography & 0.33 & 0.35 & 0.23 & 0.22 & \textbf{0.20} \\
 & \cellcolor{nstbg}\textit{all} & \cellcolor{nstbg}1.39 & \cellcolor{nstbg}1.16 & \cellcolor{nstbg}1.08 & \cellcolor{nstbg}\textbf{0.97} & \cellcolor{nstbg}1.03 \\
\midrule
\textbf{Total} & & 7.96 & 11.42 & 7.68 & \textbf{6.65} & 6.83 \\
\bottomrule
\end{tabular}}
\caption{Instance CED by error category and subtype at 10K words on GPT-5.6 Luna, in errors per 10K checked words. Rows marked \textit{all} (shaded) give the category total. The lowest value of each row is in \textbf{bold}; totals equal the instance CED of Table~\ref{tab:10k}. \NstAgent is lowest in factual detail and in 4 of the 19 subtypes, counted on unrounded values.}
\label{tab:type-10k-luna}
\end{table}

\begin{table}[htbp]
\centering
\small
\resizebox{0.918\linewidth}{!}{%
\begin{tabular}{llcccccc}
\toprule
\multirow{2}{*}{\textbf{Category}} & \multirow{2}{*}{\textbf{Subtype}} & \multicolumn{2}{c}{\textbf{20K}} & \multicolumn{2}{c}{\textbf{50K}} & \multicolumn{2}{c}{\textbf{100K}} \\
\cmidrule(lr){3-4}\cmidrule(lr){5-6}\cmidrule(lr){7-8}
 & & \textbf{RollSum} & \textbf{\NstAgent} & \textbf{RollSum} & \textbf{\NstAgent} & \textbf{RollSum} & \textbf{\NstAgent} \\
\midrule
\multirow{5}{*}{Characterization} & memory & 1.62 & \textbf{1.07} & 2.01 & \textbf{1.29} & 2.09 & \textbf{1.05} \\
 & knowledge & 0.09 & \textbf{0.06} & \textbf{0.05} & 0.06 & \textbf{0.02} & 0.08 \\
 & skill/power & \textbf{0.08} & 0.11 & 0.09 & \textbf{0.06} & \textbf{0.00} & 0.06 \\
 & forgotten ability & \textbf{0.01} & 0.02 & 0.01 & \textbf{0.00} & 0.00 & 0.00 \\
 & \cellcolor{nstbg}\textit{all} & \cellcolor{nstbg}1.79 & \cellcolor{nstbg}\textbf{1.25} & \cellcolor{nstbg}2.16 & \cellcolor{nstbg}\textbf{1.41} & \cellcolor{nstbg}2.12 & \cellcolor{nstbg}\textbf{1.19} \\
\midrule
\multirow{4}{*}{Factual detail} & appearance & 1.08 & \textbf{0.72} & 1.14 & \textbf{0.65} & 1.32 & \textbf{0.65} \\
 & nomenclature & 0.48 & \textbf{0.22} & 0.67 & \textbf{0.35} & 0.75 & \textbf{0.19} \\
 & quantitative & 1.68 & \textbf{1.58} & \textbf{1.55} & 1.74 & 1.72 & \textbf{1.38} \\
 & \cellcolor{nstbg}\textit{all} & \cellcolor{nstbg}3.23 & \cellcolor{nstbg}\textbf{2.51} & \cellcolor{nstbg}3.36 & \cellcolor{nstbg}\textbf{2.74} & \cellcolor{nstbg}3.80 & \cellcolor{nstbg}\textbf{2.22} \\
\midrule
\multirow{4}{*}{Narrative style} & perspective & 0.31 & \textbf{0.07} & 0.11 & \textbf{0.06} & 0.15 & \textbf{0.02} \\
 & tone & 0.05 & \textbf{0.02} & 0.04 & \textbf{0.00} & 0.06 & \textbf{0.00} \\
 & style shift & 0.50 & \textbf{0.22} & 0.37 & \textbf{0.20} & 0.53 & \textbf{0.23} \\
 & \cellcolor{nstbg}\textit{all} & \cellcolor{nstbg}0.86 & \cellcolor{nstbg}\textbf{0.31} & \cellcolor{nstbg}0.53 & \cellcolor{nstbg}\textbf{0.26} & \cellcolor{nstbg}0.74 & \cellcolor{nstbg}\textbf{0.25} \\
\midrule
\multirow{7}{*}{Timeline and plot} & absolute time & \textbf{0.26} & \textbf{0.26} & 0.33 & \textbf{0.31} & 0.34 & \textbf{0.21} \\
 & duration & 0.99 & \textbf{0.94} & \textbf{0.86} & 0.90 & 1.08 & \textbf{0.93} \\
 & simultaneity & 0.18 & \textbf{0.16} & 0.24 & \textbf{0.22} & 0.24 & \textbf{0.09} \\
 & causeless effect & 0.22 & \textbf{0.14} & 0.28 & \textbf{0.16} & 0.06 & \textbf{0.02} \\
 & causal logic & 0.79 & \textbf{0.65} & 0.69 & \textbf{0.57} & 0.65 & \textbf{0.32} \\
 & abandoned plot & 0.70 & \textbf{0.53} & 1.21 & \textbf{0.77} & 1.18 & \textbf{0.85} \\
 & \cellcolor{nstbg}\textit{all} & \cellcolor{nstbg}3.15 & \cellcolor{nstbg}\textbf{2.69} & \cellcolor{nstbg}3.62 & \cellcolor{nstbg}\textbf{2.93} & \cellcolor{nstbg}3.55 & \cellcolor{nstbg}\textbf{2.42} \\
\midrule
\multirow{4}{*}{World building} & core rules & \textbf{1.03} & \textbf{1.03} & \textbf{0.86} & 0.91 & 0.67 & \textbf{0.56} \\
 & social norms & 0.24 & \textbf{0.15} & 0.20 & \textbf{0.17} & 0.36 & \textbf{0.02} \\
 & geography & 0.61 & \textbf{0.41} & 0.74 & \textbf{0.50} & \textbf{0.58} & \textbf{0.58} \\
 & \cellcolor{nstbg}\textit{all} & \cellcolor{nstbg}1.88 & \cellcolor{nstbg}\textbf{1.59} & \cellcolor{nstbg}1.79 & \cellcolor{nstbg}\textbf{1.57} & \cellcolor{nstbg}1.61 & \cellcolor{nstbg}\textbf{1.16} \\
\midrule
\textbf{Total} & & 10.92 & \textbf{8.35} & 11.46 & \textbf{8.91} & 11.82 & \textbf{7.24} \\
\bottomrule
\end{tabular}}
\caption{Instance CED by error category and subtype from 20K to 100K words on DeepSeek-V4-Flash, in errors per 10K checked words. Rows marked \textit{all} (shaded) give the category total. The better method of each pair is in \textbf{bold}; totals equal the instance CED of Table~\ref{tab:100k}. \NstAgent is lower in all five categories at every length and in 17, 15, and 15 of the 19 subtypes at 20K, 50K, and 100K, counted on unrounded values.}
\label{tab:type-ds}
\end{table}

\begin{table}[htbp]
\centering
\small
\resizebox{0.918\linewidth}{!}{%
\begin{tabular}{llcccccc}
\toprule
\multirow{2}{*}{\textbf{Category}} & \multirow{2}{*}{\textbf{Subtype}} & \multicolumn{2}{c}{\textbf{20K}} & \multicolumn{2}{c}{\textbf{50K}} & \multicolumn{2}{c}{\textbf{100K}} \\
\cmidrule(lr){3-4}\cmidrule(lr){5-6}\cmidrule(lr){7-8}
 & & \textbf{RollSum} & \textbf{\NstAgent} & \textbf{RollSum} & \textbf{\NstAgent} & \textbf{RollSum} & \textbf{\NstAgent} \\
\midrule
\multirow{5}{*}{Characterization} & memory & 1.02 & \textbf{0.90} & 0.99 & \textbf{0.83} & 1.35 & \textbf{0.84} \\
 & knowledge & \textbf{0.02} & 0.05 & 0.03 & \textbf{0.01} & 0.04 & \textbf{0.02} \\
 & skill/power & \textbf{0.11} & 0.12 & 0.08 & \textbf{0.04} & 0.09 & \textbf{0.08} \\
 & forgotten ability & 0.00 & 0.00 & 0.00 & 0.00 & 0.00 & 0.00 \\
 & \cellcolor{nstbg}\textit{all} & \cellcolor{nstbg}1.14 & \cellcolor{nstbg}\textbf{1.07} & \cellcolor{nstbg}1.10 & \cellcolor{nstbg}\textbf{0.88} & \cellcolor{nstbg}1.48 & \cellcolor{nstbg}\textbf{0.94} \\
\midrule
\multirow{4}{*}{Factual detail} & appearance & 1.11 & \textbf{0.98} & \textbf{1.04} & 1.15 & \textbf{0.79} & 0.80 \\
 & nomenclature & 0.70 & \textbf{0.57} & 0.87 & \textbf{0.59} & 0.77 & \textbf{0.76} \\
 & quantitative & \textbf{0.96} & 1.01 & 0.95 & \textbf{0.89} & 1.17 & \textbf{0.70} \\
 & \cellcolor{nstbg}\textit{all} & \cellcolor{nstbg}2.76 & \cellcolor{nstbg}\textbf{2.56} & \cellcolor{nstbg}2.86 & \cellcolor{nstbg}\textbf{2.63} & \cellcolor{nstbg}2.74 & \cellcolor{nstbg}\textbf{2.26} \\
\midrule
\multirow{4}{*}{Narrative style} & perspective & 0.10 & \textbf{0.09} & 0.06 & \textbf{0.05} & \textbf{0.00} & 0.02 \\
 & tone & 0.01 & \textbf{0.00} & 0.00 & 0.00 & 0.00 & 0.00 \\
 & style shift & 0.26 & \textbf{0.06} & 0.28 & \textbf{0.08} & 0.30 & \textbf{0.20} \\
 & \cellcolor{nstbg}\textit{all} & \cellcolor{nstbg}0.38 & \cellcolor{nstbg}\textbf{0.14} & \cellcolor{nstbg}0.34 & \cellcolor{nstbg}\textbf{0.13} & \cellcolor{nstbg}0.30 & \cellcolor{nstbg}\textbf{0.22} \\
\midrule
\multirow{7}{*}{Timeline and plot} & absolute time & \textbf{0.21} & 0.23 & 0.25 & \textbf{0.10} & 0.27 & \textbf{0.17} \\
 & duration & 0.57 & \textbf{0.53} & 0.51 & \textbf{0.42} & 0.54 & \textbf{0.31} \\
 & simultaneity & 0.26 & \textbf{0.19} & \textbf{0.16} & 0.17 & \textbf{0.10} & 0.15 \\
 & causeless effect & 0.47 & \textbf{0.25} & \textbf{0.24} & 0.33 & 0.18 & \textbf{0.13} \\
 & causal logic & 0.51 & \textbf{0.42} & 0.51 & \textbf{0.43} & 0.52 & \textbf{0.30} \\
 & abandoned plot & 1.19 & \textbf{1.06} & 1.67 & \textbf{1.55} & 1.94 & \textbf{1.90} \\
 & \cellcolor{nstbg}\textit{all} & \cellcolor{nstbg}3.20 & \cellcolor{nstbg}\textbf{2.67} & \cellcolor{nstbg}3.34 & \cellcolor{nstbg}\textbf{3.00} & \cellcolor{nstbg}3.54 & \cellcolor{nstbg}\textbf{2.97} \\
\midrule
\multirow{4}{*}{World building} & core rules & 0.87 & \textbf{0.71} & 0.73 & \textbf{0.57} & 0.55 & \textbf{0.33} \\
 & social norms & \textbf{0.22} & 0.24 & 0.32 & \textbf{0.14} & 0.26 & \textbf{0.15} \\
 & geography & 0.40 & \textbf{0.31} & 0.40 & \textbf{0.34} & 0.48 & \textbf{0.35} \\
 & \cellcolor{nstbg}\textit{all} & \cellcolor{nstbg}1.48 & \cellcolor{nstbg}\textbf{1.25} & \cellcolor{nstbg}1.45 & \cellcolor{nstbg}\textbf{1.05} & \cellcolor{nstbg}1.30 & \cellcolor{nstbg}\textbf{0.83} \\
\midrule
\textbf{Total} & & 8.97 & \textbf{7.70} & 9.10 & \textbf{7.69} & 9.37 & \textbf{7.22} \\
\bottomrule
\end{tabular}}
\caption{Instance CED by error category and subtype from 20K to 100K words on GPT-5.6 Luna, in errors per 10K checked words. Rows marked \textit{all} (shaded) give the category total. The better method of each pair is in \textbf{bold}; totals equal the instance CED of Table~\ref{tab:100k}. \NstAgent is lower in all five categories at every length and in 13, 14, and 14 of the 19 subtypes at 20K, 50K, and 100K, counted on unrounded values.}
\label{tab:type-luna}
\end{table}

\begin{table}[htbp]
\centering
\small
\resizebox{0.767\linewidth}{!}{%
\begin{tabular}{llcccc}
\toprule
\textbf{Category} & \textbf{Subtype} & \textbf{\NstAgent} & \textbf{$-$State} & \textbf{$-$Lookback} & \textbf{RollSum} \\
\midrule
\multirow{5}{*}{Characterization} & memory & \textbf{1.07} & 1.49 & 1.40 & 1.62 \\
 & knowledge & 0.06 & 0.06 & \textbf{0.05} & 0.09 \\
 & skill/power & 0.11 & \textbf{0.06} & 0.08 & 0.08 \\
 & forgotten ability & 0.02 & \textbf{0.00} & \textbf{0.00} & 0.01 \\
 & \cellcolor{nstbg}\textit{all} & \cellcolor{nstbg}\textbf{1.25} & \cellcolor{nstbg}1.61 & \cellcolor{nstbg}1.52 & \cellcolor{nstbg}1.79 \\
\midrule
\multirow{4}{*}{Factual detail} & appearance & 0.72 & \textbf{0.64} & 0.72 & 1.08 \\
 & nomenclature & \textbf{0.22} & 0.43 & 0.48 & 0.48 \\
 & quantitative & 1.58 & \textbf{1.56} & 1.69 & 1.68 \\
 & \cellcolor{nstbg}\textit{all} & \cellcolor{nstbg}\textbf{2.51} & \cellcolor{nstbg}2.62 & \cellcolor{nstbg}2.89 & \cellcolor{nstbg}3.23 \\
\midrule
\multirow{4}{*}{Narrative style} & perspective & 0.07 & \textbf{0.06} & 0.17 & 0.31 \\
 & tone & \textbf{0.02} & \textbf{0.02} & 0.04 & 0.05 \\
 & style shift & \textbf{0.22} & 0.35 & 0.29 & 0.50 \\
 & \cellcolor{nstbg}\textit{all} & \cellcolor{nstbg}\textbf{0.31} & \cellcolor{nstbg}0.43 & \cellcolor{nstbg}0.50 & \cellcolor{nstbg}0.86 \\
\midrule
\multirow{7}{*}{Timeline and plot} & absolute time & 0.26 & \textbf{0.25} & 0.36 & 0.26 \\
 & duration & 0.94 & \textbf{0.93} & 1.10 & 0.99 \\
 & simultaneity & 0.16 & \textbf{0.15} & 0.17 & 0.18 \\
 & causeless effect & \textbf{0.14} & 0.23 & 0.23 & 0.22 \\
 & causal logic & \textbf{0.65} & 0.83 & 0.68 & 0.79 \\
 & abandoned plot & \textbf{0.53} & 0.84 & 0.62 & 0.70 \\
 & \cellcolor{nstbg}\textit{all} & \cellcolor{nstbg}\textbf{2.69} & \cellcolor{nstbg}3.24 & \cellcolor{nstbg}3.16 & \cellcolor{nstbg}3.15 \\
\midrule
\multirow{4}{*}{World building} & core rules & 1.03 & 0.93 & \textbf{0.92} & 1.03 \\
 & social norms & \textbf{0.15} & 0.16 & 0.16 & 0.24 \\
 & geography & \textbf{0.41} & 0.42 & 0.64 & 0.61 \\
 & \cellcolor{nstbg}\textit{all} & \cellcolor{nstbg}1.59 & \cellcolor{nstbg}\textbf{1.51} & \cellcolor{nstbg}1.71 & \cellcolor{nstbg}1.88 \\
\midrule
\textbf{Total} & & \textbf{8.35} & 9.40 & 9.79 & 10.92 \\
\bottomrule
\end{tabular}}
\caption{Instance CED by error category and subtype for the ablations of Table~\ref{tab:ablation} (DeepSeek-V4-Flash, 20K words), in errors per 10K checked words. Rows marked \textit{all} (shaded) give the category total. The lowest value of each row is in \textbf{bold}; totals equal the instance CED of Table~\ref{tab:ablation}. \NstAgent is lowest in four of the five categories and in 9 of the 19 subtypes, counted on unrounded values.}
\label{tab:type-ablation}
\end{table}

\FloatBarrier
\subsection{A Final State at 100K Words}
\label{app:state-example}

The state is the agent's entire memory of what it has written, so its final contents are worth
inspecting directly. The example below is the end state of one 100K-word story on DeepSeek-V4-Flash,
abridged to one entry of each kind. It holds 17 character snapshots, 153 past events, and 6 unresolved
requirements, about 18.5K words in total against the 110.7K words of the story it describes. This is
what each chapter prompt carries in place of the prose: a sixth of the words written so far.

\begin{promptbox}{Final narrative state of one 100K-word story (one entry of each type)}
\textbf{Character state.} \emph{Silas Vane.} Memoryless vessel now returned to the surface, tending the
Gilt Ledger counter once more. Emerged from the west breach of the Aethelmarrow at dawn (the day after
the fall, 27th of Tallow), unrecognizable even to himself, guided up by his hands' knowledge of the
counter. At the anchor-head he unbraided Marten's forty-one-year grey rope from his belt \dots

\textbf{Past event.} Silas Vane bought a memory-keepsake from grieving widow Calla Ostergaard: a small
sealed glass vessel containing a wisp of grey light that is actually a stolen fragment of a dead
ancestor's singing --- a piece of a life cut out and kept alive, not a simple recording. Silas
recognized it as a debt rather than a trinket \dots

\textbf{Open future requirement.} Delphine Mallory, first-tier widow, had her two winters of grief
drained from her by Crowe's ringing of the Bell; she walked away lighter but hollowed out, grateful and
unaware anything was taken. She is the first of the Bell's hollowed-out victims; Crowe's appetite will
seek more \dots
\end{promptbox}

Two properties of this state are worth noting against a free-text summary of the same story. Each entry
is addressed by a stable key or character name, so an update rewrites one entry and leaves the rest
byte-identical, and the open requirements are a list the agent can enumerate at any point rather than a
promise implicitly encoded in prose. Both are what make the failure mode in Appendix~\ref{app:case}
visible: when a fact is wrong in the state, it is wrong in one identifiable place.

\subsection{Baselines Limited to 10K Words}
\label{app:baseline-scaling}

Three of the four baselines are evaluated only at 10K words. The reason is not that they score poorly
at longer targets but that their released designs do not define a run at those targets, and forcing one
would compare our reimplementation rather than the published method.

\emph{Direct} must emit the story in one call. At 10K words this already fails often
(Section~\ref{sec:challenge}); at a 20K target the lower bound of the acceptance band is 16K words, which leaves little
or no room within the 32,768-token output limit once reasoning tokens are counted. Extending it is a question of training
longer-writing models~\citep{bai2025longwriter,wu2026longwriterzero,quan2024selflengthen}, not of
changing the workflow.

\emph{DOME}~\citep{wang2025dome} fixes the story to five acts and maintains a knowledge graph with
per-triple LLM calls. The act count is structural: reaching 100K words within five acts requires
chapters of about 20K words each, which no backbone can write in one call under the length gate. The
graph is also the dominant cost already at 10K, about 3{,}500 calls per story
(Table~\ref{tab:efficiency}).

\emph{StoryWriter}~\citep{xia2025storywriter} plans five to ten events with three sub-events each, so
its released configuration spans a comparable range; scaling it to 100K words means changing the event
schedule, which is a different method from the published one.

RollSum is the one baseline whose design is length-agnostic: it rewrites a summary after every chapter
and therefore runs unchanged at any target. That is why the scaling comparison is against RollSum, and
why we treat it as the representative of free-text memory rather than as a weak baseline: at 10K words
it is the strongest baseline on both backbones, and on GPT-5.6 Luna it matches \NstAgent. Other recent
long-form systems make the same structural commitments we describe here, whether recursive
planning~\citep{xiong2025writehere}, extraction-and-expansion~\citep{huang2024ex3}, multi-agent
collaboration~\citep{huot2025agentsroom,venkatraman2025collabstory,yu2025charactersimulation}, or
graph-structured plots~\citep{gu2026plotter,shi2025kgtheory}; a controlled comparison across an order of
magnitude of length requires a baseline that is defined at every length.

\subsection{Statistical Analysis}
\label{app:stats}

Every number in Tables~\ref{tab:10k},~\ref{tab:100k}, and~\ref{tab:ablation} is a mean over per-story
scores, and every comparison we discuss is paired: the two methods write from the same prompt and the
same frozen outline, so we test the per-prompt difference with a paired $t$-test, a 95\% bootstrap
confidence interval (10{,}000 resamples), and a Holm correction applied within each metric, that is,
across the comparisons a reader scans in one column. Dispersion is summarized by the standard error of
the mean; at 10K words the per-story standard deviation of instance CED is 3.3 for \NstAgent and 4.3
for RollSum on DeepSeek-V4-Flash, so differences below roughly 0.7 are within noise at $n=100$. We mark
as significant only what survives these corrections, and every table reports one evaluation per arm.

At 10K words (Table~\ref{tab:10k}) on DeepSeek-V4-Flash, \NstAgent improves instance CED by $-2.78$
[$-3.65$, $-1.95$] over RollSum, $-2.93$ [$-3.99$, $-1.89$] over Direct, and $-5.40$ [$-6.69$,
$-4.17$] over StoryWriter, all $p<0.001$ after correction; on GPT-5.6 Luna only the instance-CED gap over Direct
survives the correction ($-1.28$ [$-2.21$, $-0.33$], $p=0.024$), together with the writing-quality gap
over StoryWriter ($+0.85$ [$+0.66$, $+1.06$], $p<0.001$); against RollSum the difference is flat. DOME is excluded from these tests because it was run on a different 20-prompt set.
Table~\ref{tab:stats-scaling} gives the comparison with RollSum at every length, and
Table~\ref{tab:ablation-stats} the ablations of Table~\ref{tab:ablation}.

\begin{table}[h]
\centering
\small
\resizebox{\linewidth}{!}{%
\begin{tabular}{llcccc}
\toprule
\textbf{Model} & \textbf{Length} & \textbf{Subtype CED} & \textbf{Instance CED} & \textbf{Local-only Instance CED} & \textbf{Writing Quality} \\
\midrule
\multirow{4}{*}{DeepSeek-V4-Flash}
 & 10K & $-1.44$ [$-1.86$, $-1.03$]$^{***}$ & $-2.78$ [$-3.65$, $-1.95$]$^{***}$ & $-2.42$ [$-3.28$, $-1.61$]$^{***}$ & $+0.05$ [$-0.05$, $+0.15$] \\
 & 20K & $-1.38$ [$-1.83$, $-0.94$]$^{***}$ & $-2.57$ [$-3.52$, $-1.63$]$^{***}$ & $-2.40$ [$-3.31$, $-1.48$]$^{***}$ & $+0.15$ [$+0.05$, $+0.26$]$^{*}$ \\
 & 50K & $-1.43$ [$-1.91$, $-0.96$]$^{***}$ & $-2.59$ [$-3.58$, $-1.61$]$^{***}$ & $-2.15$ [$-3.10$, $-1.22$]$^{***}$ & $+0.22$ [$+0.14$, $+0.29$]$^{***}$ \\
 & 100K & $-2.38$ [$-3.20$, $-1.59$]$^{***}$ & $-4.58$ [$-6.37$, $-2.85$]$^{***}$ & $-4.25$ [$-6.01$, $-2.54$]$^{***}$ & $+0.24$ [$+0.09$, $+0.40$]$^{*}$ \\
\midrule
\multirow{4}{*}{GPT-5.6 Luna}
 & 10K & $+0.21$ [$-0.30$, $+0.70$] & $+0.17$ [$-0.69$, $+1.01$] & $+0.01$ [$-0.85$, $+0.84$] & $+0.03$ [$-0.05$, $+0.11$] \\
 & 20K & $-0.86$ [$-1.38$, $-0.35$]$^{**}$ & $-1.27$ [$-2.11$, $-0.44$]$^{*}$ & $-1.14$ [$-1.93$, $-0.34$]$^{*}$ & $+0.05$ [$-0.01$, $+0.11$] \\
 & 50K & $-0.74$ [$-1.27$, $-0.22$]$^{*}$ & $-1.40$ [$-2.34$, $-0.49$]$^{*}$ & $-1.28$ [$-2.19$, $-0.38$]$^{*}$ & $+0.01$ [$-0.05$, $+0.09$] \\
 & 100K & $-0.88$ [$-1.68$, $-0.06$] & $-2.15$ [$-3.47$, $-0.85$]$^{*}$ & $-2.12$ [$-3.43$, $-0.81$]$^{*}$ & $+0.18$ [$+0.10$, $+0.26$]$^{***}$ \\
\bottomrule
\end{tabular}}
\caption{Paired differences between \NstAgent and RollSum (\NstAgent minus RollSum) at each length,
with 95\% bootstrap confidence intervals. $^{*}$, $^{**}$, $^{***}$ mark $p<0.05$, $p<0.01$, $p<0.001$
after Holm correction within each metric, over the four lengths in that column of a backbone's block. \emph{Local-only instance CED} excludes the globally scoped
abandoned\_plot\_elements subtype; see the discussion of Table~\ref{tab:global-local}.}
\label{tab:stats-scaling}
\end{table}

\begin{table}[h]
\centering
\small
\resizebox{\linewidth}{!}{%
\begin{tabular}{lccc}
\toprule
\textbf{Variant $-$ \NstAgent} & \textbf{Subtype CED} & \textbf{Instance CED} & \textbf{Writing Quality} \\
\midrule
$-$State & $+0.52$ [$+0.10$, $+0.94$]$^{*}$ & $+1.05$ [$+0.16$, $+1.95$]$^{*}$ & $-0.09$ [$-0.17$, $-0.02$]$^{*}$ \\
$-$Lookback & $+1.20$ [$+0.74$, $+1.65$]$^{***}$ & $+1.44$ [$+0.61$, $+2.23$]$^{**}$ & $-0.16$ [$-0.23$, $-0.09$]$^{***}$ \\
RollSum & $+1.38$ [$+0.94$, $+1.83$]$^{***}$ & $+2.57$ [$+1.63$, $+3.52$]$^{***}$ & $-0.15$ [$-0.26$, $-0.05$]$^{*}$ \\
\bottomrule
\end{tabular}}
\caption{Paired differences for Table~\ref{tab:ablation} (variant minus the full agent, so positive CED
means the variant is worse), with 95\% bootstrap confidence intervals and Holm correction within each
metric, over the three comparisons in that column. Removing either memory channel makes the agent
measurably worse on all three metrics.}
\label{tab:ablation-stats}
\end{table}

\paragraph{The same 50 prompts at every length.}
The 100K setting uses 50 of the 100 prompts, so the length trend could in principle reflect which
prompts were chosen. Because those 50 are a subset of the 100 used at the shorter lengths, we recompute
every length on exactly that subset. Table~\ref{tab:subset} shows the resulting gap between \NstAgent
and RollSum. The pattern of Table~\ref{tab:100k} survives: on DeepSeek-V4-Flash the gap in instance CED
is stable from 10K to 50K and widens at 100K, and on GPT-5.6 Luna it starts near zero at 10K and grows
with length. Sample composition therefore does not explain the trend.

\begin{table}[h]
\centering
\small
\resizebox{\linewidth}{!}{%
\begin{tabular}{lcccc}
\toprule
\textbf{Model} & \textbf{10K} & \textbf{20K} & \textbf{50K} & \textbf{100K} \\
\midrule
DeepSeek-V4-Flash & $-2.60$ [$-3.89$, $-1.32$] & $-2.48$ [$-3.74$, $-1.24$] & $-2.76$ [$-4.13$, $-1.37$] & $-4.58$ [$-6.37$, $-2.85$] \\
GPT-5.6 Luna & $+0.48$ [$-0.75$, $+1.65$] & $-1.29$ [$-2.48$, $-0.12$] & $-2.03$ [$-3.26$, $-0.83$] & $-2.15$ [$-3.47$, $-0.85$] \\
\bottomrule
\end{tabular}}
\caption{Instance CED difference (\NstAgent minus RollSum, negative favors \NstAgent) computed on the
same 50 prompts at every length, with 95\% bootstrap confidence intervals. $n=50$ except
DeepSeek-V4-Flash at 50K, where RollSum completed 49 of the 50.}
\label{tab:subset}
\end{table}

\section{Reinforcement Learning for \texorpdfstring{\NstAgent}{NstAgent}}
\label{app:rl}

\paragraph{Setup.}
We examine whether the full \NstAgent chapter loop can be optimized with reinforcement learning, starting from Qwen3.5-4B with reasoning enabled. Training uses 256 English premises (32 genres, eight each) that do not overlap with the ConStory-Bench prompts, each planned at 10K, 20K, 50K, and 100K words. For every story, the first nine chapters are prepared with the initial model and saved as prefixes, and
the policy writes the next chapter from a sampled prefix index $j\in\{0,\dots,9\}$ through the same
tools and state update as \NstAgent. Prefixes are audited before training so that every sampled context
carries a well-formed state and a chapter plan; at $j=0$ there is no prior chapter, and the consistency
judge is told to score internal coherence against the chapter plan instead of inventing a history. The
rollout uses the deployed agent loop rather than a simplified environment, so a training episode can
fail exactly as generation does, by exhausting the turn budget, by never passing the length gate, or by
submitting an update the schema rejects.

\paragraph{Training and reward.}
We run chapter-level GRPO~\citep{shao2024deepseekmath} with a group size of 8 and 32 distinct contexts per step, with learning rate $5\times10^{-7}$ and KL coefficient 0.01; tool-response tokens are masked from the loss. We report checkpoints up to step 48. A rollout receives zero reward unless it completes a successful \textbf{write}, one successful \textbf{update}, and \textbf{DONE}. Otherwise the reward is
\begin{equation}
R = \mathbb{1}\big[\,|c|\in[0.8w,\ 1.2w]\,\big]\cdot\Big(\tfrac{1}{2}\cdot\tfrac{C+Q}{2} + \tfrac{1}{2}S\Big),
\end{equation}
where $C$, $Q$, and $S$ are scores in $[0,1]$ from separate DeepSeek-V4-Flash judges for the new
chapter's consistency with prior chapters, its writing quality, and the fidelity of the state update.

\paragraph{Reward judges.}
The three judges share a preamble and differ only in their rubric, which keeps the scales comparable.
The preamble states that the supplied JSON is untrusted story data rather than instructions, fixes the
target of the score (the new chapter, or for $S$ the submitted update), forbids penalizing an earlier
chapter's independent flaws unless the new chapter repeats them, and requires the score to be returned
as a bare JSON object without a reason or quotation; the consistency judge first reports whether the new
chapter contains an explicit same-time factual contradiction. Free-text evidence is omitted to bound the
cost of judging long chapters.
The consistency rubric adds explicit anchors: 9--10 for no supported continuity defect, 5--6 for a
consequential unexplained change, and at most 4 when the chapter contains an explicit, unqualified
contradiction, with the instruction not to describe a contradiction in the evidence and then award a
high score. It also lists what is \emph{not} a contradiction, namely quoted lies, disputed testimony,
dreams, figurative language, and established supernatural rules, and it excludes prose repetition,
pacing, and literary impact, which belong to the quality judge. 

\paragraph{The state rubric.}
The state update is scored as a transition: whether upserts preserve still-relevant fields while incorporating real changes, whether
past events actually occurred and add information not already explicit in the frozen outline, whether
requirements are concrete unresolved obligations, and whether resolutions follow a supported payoff. The rubric
penalizes fabricated facts, unjustified resolution, lost persistent facts, and redundant copying of the
outline, and it treats the executed post-state as evidence rather than as ground truth, so a failed
update cannot earn credit for the correct update it intended. Judges run with a 32{,}768-token output
limit and a 900-second timeout at concurrency 8; a malformed or missing score makes the rollout score
zero for that metric rather than being retried into the reward.

\paragraph{Evaluation.}
We evaluate the initial model and checkpoints at steps 16, 32, and 48 on 20 ConStory-Bench prompts at 10K words, with outlines generated once by the initial model and shared by all checkpoints. Stories are judged by DeepSeek-V4-Pro with WritingBench and extended ConStory-Bench with every chapter marked.

\begin{table}[h]
\centering
\small
\begin{tabular}{lccc}
\toprule
\textbf{Checkpoint} & \textbf{Subtype CED ($\downarrow$)} & \textbf{Instance CED ($\downarrow$)} & \textbf{Writing Quality ($\uparrow$)} \\
\midrule
Initial (Qwen3.5-4B) & 9.617 & 15.627 & 4.80 \\
Step 16 & 9.471 & 16.756 & 4.60 \\
Step 32 & 9.263 & 14.373 & 4.86 \\
Step 48 & \textbf{7.672} & \textbf{12.135} & \textbf{5.20} \\
\bottomrule
\end{tabular}
\caption{RL checkpoints of \NstAgent on Qwen3.5-4B at 10K words (20 prompts, shared outlines). Best in \textbf{bold}.}
\label{tab:rl}
\end{table}

\paragraph{Results.}
Table~\ref{tab:rl} shows little change during the first 32 steps. By step 48 the policy writes chapters that the judges rate higher in quality and that contain clearly fewer contradictions, with instance-level error density about a fifth lower than the initial model's. These results indicate that the state-tracking loop provides a usable training signal even for a small open-weight model. They come from one model, one seed, and 20 prompts that were also used to compare checkpoints, and we do not claim improvements at longer lengths.

\section{Case Studies}
\label{app:case}

We examine three story pairs in which \NstAgent and RollSum write from the same prompt and the same frozen outline: two where explicit state prevents an error that free-text memory makes, and one where it produces an error that free-text memory avoids. Contradictions were flagged by the extended ConStory-Bench judge, and we checked each quoted passage against the generated text. Chapters are numbered from 0, following the evaluation.

\subsection{Case 1: Repeated Events (DeepSeek-V4-Flash, 20K)}

\paragraph{Setting.}
The prompt asks for a world of six elemental kingdoms shattered by a slumbering titan, where heroes from each kingdom must unite. In both stories, the protagonist Kaelen carries a prophecy-stone that sings whenever it meets another of the six chosen heroes, so the same kind of event recurs throughout the book.

\paragraph{RollSum.}
Table~\ref{tab:case1} traces how the contradiction arises. The stone sings for Nerys in Chapter 2 and for Cyra in Chapter 3. By Chapter 8, the summary has paraphrased these moments as the stone ``pulsing'' and ``humming'', and records that it ``has never sung for him''. Writing from this summary, Chapter 8 announces that the stone sings for the first time since the Ember-Scar, which contradicts both earlier chapters. The summary is 2,644 words long at this point, so the loss comes from paraphrase during repeated rewriting rather than from a lack of space.

\paragraph{\NstAgent.}
\NstAgent records each occurrence as a separate keyed past event at the time it happens, so the state that accompanies every later chapter still says that the stone sang for Nerys and for Cyra. It also turned the stone's behavior into a future requirement in Chapter 1 and resolved it in Chapter 2. The judge flags a single contradiction in \NstAgent's last seven chapters, compared with ten in RollSum's last eight. The prospective part of the state behaves similarly: all sixteen requirements \NstAgent created were resolved by the final chapter, whereas RollSum's antagonists vow in Chapter 7 to wait ``at every facet'' and never reappear, which the judge flags as an abandoned plot element.

\begin{table}[h]
\centering
\small
\renewcommand{\arraystretch}{1.15}
\begin{tabular}{>{\raggedright\arraybackslash}p{0.25\linewidth}>{\raggedright\arraybackslash}p{0.69\linewidth}}
\toprule
\textbf{Source} & \textbf{Excerpt} \\
\midrule
RollSum, Chapter 2 & ``He drew it out. The runes shifted like sand in a current, and the stone sang---a hum that answered something deep in the water, in the coral, in her.'' \\
RollSum, Chapter 3 & ``The stone sang---a thin, high note, a string that had been waiting to be plucked.'' \\
RollSum summary before Chapter 8 & ``The stone has never sung for him; it only pulses warmly.'' \ldots\ ``The prophecy-stone hums in the braided voices of the found souls \ldots'' \\
RollSum, Chapter 8 & ``The prophecy-stone at Kaelen's hip pulsed, warm as a coal, and for the first time since the Ember-Scar it began to sing.'' \\
\midrule
\oursbg \NstAgent state, past event \texttt{nerys\_joins\_kaelen} & \oursbg ``On the drowned Tidal spire, Kaelen found Nerys \ldots\ The prophecy-stone sang in her presence \ldots'' \\
\oursbg \NstAgent state, past event \texttt{cyra\_joins\_kaelen} & \oursbg ``At the Gyre Bastion \ldots\ the prophecy-stone sang for Cyra, the disgraced wind-rider.'' \\
\oursbg \NstAgent state, requirement \texttt{prophecy\_\allowbreak stone\_\allowbreak responds\_\allowbreak to\_\allowbreak champions} & \oursbg Added in Chapter 1: ``The prophecy-stone \ldots\ must be shown to respond to the presence of other unbounded souls \ldots''; resolved in Chapter 2. \\
\bottomrule
\end{tabular}
\caption{Case 1. The rolling summary rewrites earlier singing as humming, and the next chapter declares a ``first time''. \NstAgent keeps each occurrence as a keyed past event.}
\label{tab:case1}
\end{table}

\subsection{Case 2: Identity Facts over 40 Chapters (GPT-5.6 Luna, 100K)}

\paragraph{Setting.}
The prompt describes a VHS tape that implants false childhood memories in anyone born after 1990, and asks for containment procedures, breach logs, and testimonies. Birth dates are therefore plot-critical: they decide who may approach the tape, and characters repeat them as identity checks throughout the story.

\paragraph{RollSum.}
Table~\ref{tab:case2} shows that RollSum's errors follow a consistent path. A birth date is stated precisely in an early chapter, drifts when the summary is regenerated, and the drifted value is then written into the final chapters. Dr.~Vale is born in 1961 in Chapter 13 but ``approximately 1963'' in the summary and on 6 February 1963 in Chapter 38. Daniel's personnel file gives March 14, 1994 in Chapter 2, while the summary and Chapter 36 give October 17. Mara gives her date of birth as July 1987 in Chapter 6, but June 1988 from Chapter 7 onward, and the summary carries 1988 forward to the end. The summary had grown to 4,278 words by Chapter 36, yet it still did not preserve these values verbatim. The judge flags 25 contradictions in RollSum's last four chapters, most of them in birth dates, ages, and personal histories.

\paragraph{\NstAgent.}
\NstAgent stores these facts in character snapshots, which are rewritten in full at every update. Once Dr.~Vale states that she was born in 1969 (Chapter 24), each of her subsequent snapshots, from Chapter 28 through Chapter 38, repeats that year. Owen Bell's snapshot records 1993 from Chapter 12 to the end. Daniel's birth year stays 1993 across Chapters 0, 1, 7, 28, and 29. The judge flags 10 contradictions in \NstAgent's last four chapters.

\begin{table}[h]
\centering
\small
\renewcommand{\arraystretch}{1.15}
\begin{tabular}{>{\raggedright\arraybackslash}p{0.25\linewidth}>{\raggedright\arraybackslash}p{0.69\linewidth}}
\toprule
\textbf{Source} & \textbf{Excerpt} \\
\midrule
RollSum, Chapter 13 & Dr.~Vale: ``I was born in 1961.'' \\
RollSum summary before Chapter 36 & ``Vale, born approximately 1963, spent summers at a farmhouse with her younger sister Miriam \ldots'' \\
RollSum, Chapter 38 & ``Irena Vale. Born 6 February 1963.'' \\
\midrule
RollSum, Chapter 2 & ``Daniel Reyes, born March 14, 1994, Newark, New Jersey.'' \\
RollSum summary before Chapter 36 & ``Daniel, born 17 October 1994, directly viewed the tape \ldots'' \\
RollSum, Chapter 36 & ``You're Daniel Mateo Reyes. Born October seventeenth, nineteen ninety-four.'' \\
\midrule
\oursbg \NstAgent, Chapter 24 & \oursbg Dr.~Vale: ``I had been born in 1969, so the date was possible.'' \\
\oursbg \NstAgent state, Vale snapshot (Chapters 28--38) & \oursbg ``\ldots\ Vale remains active, exhausted, and factually oriented as a pre-cutoff director born in 1969.'' \\
\oursbg \NstAgent state, Owen snapshot (Chapters 12--38) & \oursbg ``\ldots\ barred from the restricted media room because he was born in 1993, after the January 1, 1991 vulnerability cutoff \ldots'' \\
\bottomrule
\end{tabular}
\caption{Case 2. In RollSum, birth dates drift when the summary is regenerated and the drifted values reach the final chapters. In \NstAgent, character snapshots restate the recorded year at every update.}
\label{tab:case2}
\end{table}

\subsection{Case 3: An Error Fixed in the State (GPT-5.6 Luna, 10K)}

\paragraph{Setting.}
The prompt asks for a horror story about a discredited priest investigating killings in a seaside
asylum during a storm. Both stories give the priest a superior who ordered an earlier exorcism, and a
victim, Mara, whose age at that exorcism is stated in the opening chapter and recalled later. The same
mechanism that preserves a correct fact preserves an incorrect one, so we looked for the stories where
\NstAgent does worst relative to RollSum: at 10K words on GPT-5.6 Luna, the largest gap in the wrong
direction is this pair, with 15 contradictions flagged for \NstAgent against 6 for RollSum.

\paragraph{RollSum.}
Table~\ref{tab:case3} shows that RollSum keeps both facts fixed. It calls the superior Bishop Creel in
Chapter 0 and in four later chapters, and states Mara's age as nine in both chapters that mention it.
Its memory is not what prevents the error: its own chapters never rename anyone, so there is nothing
for the summary to preserve or lose.

\paragraph{\NstAgent.}
\NstAgent names the same character Bishop Armitage in Chapter 0 and Bishop Haldane from Chapter 4
onward. Armitage never appears again after Chapter 0, Haldane appears in five of the remaining nine
chapters, and the state records a past event that names him Haldane, so every later chapter is written
against the new name. The same story also drifts on Mara's age: she is nine when brought to the rectory
in Chapter 0 and eight in a Chapter 6 flashback. The judge flags the rename as both a memory
contradiction and a nomenclature confusion.

\begin{table}[h]
\centering
\small
\renewcommand{\arraystretch}{1.15}
\begin{tabular}{>{\raggedright\arraybackslash}p{0.25\linewidth}>{\raggedright\arraybackslash}p{0.69\linewidth}}
\toprule
\textbf{Source} & \textbf{Excerpt} \\
\midrule
RollSum, Chapter 0 & ``Behind him, Bishop Creel had said, `Continue.' '' \\
RollSum, Chapter 8 & ``For one breath it was Bishop Creel, broad and pale beneath his red skullcap.'' \\
RollSum, Chapters 0 and 4 & ``Mara had been nine, narrow-shouldered and solemn \ldots'' \ldots\ ``\ldots\ nine years old, wrists bound, candles guttering whenever she breathed.'' \\
\midrule
\oursbg \NstAgent, Chapter 0 & \oursbg ``Bishop Armitage ordered Elias to conduct the exorcism, though he had been ordained only two years and had never performed one alone.'' \\
\oursbg \NstAgent, Chapter 4 & \oursbg ``Bishop Haldane said the demon would imitate her, and that hesitation would damn her.'' \\
\oursbg \NstAgent state, past event & \oursbg ``Elias confessed that Mara had asked him to stop during her exorcism \ldots\ because he feared disobeying Bishop Haldane more than losing her.'' \\
\oursbg \NstAgent, Chapters 0 and 6 & \oursbg ``Mara Venn had been nine when they brought her to Saint Bartholomew's rectory.'' \ldots\ ``Mara stepped through, eight years old, her nightdress dark with sweat.'' \\
\bottomrule
\end{tabular}
\caption{Case 3. \NstAgent renames the bishop in Chapter 4 and records the new name in the state, after
which every later chapter uses it; RollSum keeps one name and one age throughout.}
\label{tab:case3}
\end{table}

\paragraph{Why the state does not prevent this.}
The state is written by the same model that writes the prose, from the chapter it has just written. If
that chapter renames a character, the update faithfully records the new name, and every subsequent
chapter is then conditioned on the wrong value. Nothing in the loop compares a new entry against the
earlier text: \textbf{read} and \textbf{search} exist for exactly this check, but the writer invokes
them only when it decides to, and GPT-5.6 Luna almost never searches (Table~\ref{tab:tools}).

\paragraph{What the three cases show together.}
Explicit state converts one class of error into another: it removes the drift of repeated paraphrase
(Cases 1 and 2) but locks in a fact extracted incorrectly (Case 3), and the \textbf{correct} tool that
could repair it is invoked 0.1--2.3 times per story (Table~\ref{tab:tools}). A verification step that checks new
state entries against the source chapters, rather than trusting the writer's extraction, is the obvious
next design and is not part of the present system; post-hoc rewriting~\citep{cui2026storylens} and
coherence checks on a finished draft~\citep{zhang2025mldea} attack the same errors from the other end,
and evaluations of whether a character stays in role~\citep{spotting2025ooc,song2026arcane} would detect
this particular failure. We report this case because the aggregate numbers,
which favor \NstAgent at every length beyond 10K, do not show it.

\section{Limitations}
\label{app:limitations}

The benefit of \NstAgent depends on story length and backbone: for short stories written by a strong
model, a free-text summary performs comparably. All experiments use English prompts and no human
evaluation, so our conclusions concern the metrics rather than reader
experience~\citep{mei2026illusions,rashkin2025feedback}, and CED does not capture other failures of long
fiction such as stylistic flattening~\citep{dk2026samestories}, repeated
phrasing~\citep{tanakaishii2026repeated}, or assembly from unrelated
fragments~\citep{pham2026frankentext,ma2026textsurvey,teleki2025llmsurvey,liu2026narrativetheorysurvey,park2026consistencysurvey}.
The baselines are adaptations: DOME and StoryWriter may not reflect every design choice of their released
implementations, and RollSum isolates passive summarization rather than dual-memory designs such as
RecurrentGPT~\citep{zhou2023recurrentgpt}.

Consistency is scored by an LLM judge whose precision is imperfect. In the injection study of
Appendix~\ref{app:judge-recall}, recall does not decay with prefix length, but the judge reports a
contradiction of the targeted type for 25\% of control passages that contain none, and files a single
injected contradiction under about four subtypes. Instance CED therefore counts one underlying error more
than once, and Subtype CED is the more conservative of the two measures. The judge, DeepSeek-V4-Pro, also
shares a model family with the DeepSeek-V4-Flash backbone, and judges can favor outputs of their own
family~\citep{panickssery2024selfpreference}. We therefore read ConStory-Bench as an
evidence-grounded diagnostic rather than an exact count of errors.

The near-linear cost of Table~\ref{tab:efficiency} depends on prefix caching. Past events accumulate
without eviction, so input grows faster than length: 7.5M input tokens per story at 100K words against
0.71M at 10K, of which about 60\% is served from the provider's cache. Priced entirely as uncached input,
a 100K-word story would cost USD 2.28 instead of 1.32. Bounding or consolidating the past-event log is
left for future work.

\end{document}